\documentclass[utf8]{frontiersSCNS} 

\usepackage{url, hyperref, lineno, microtype, subcaption}
\usepackage[onehalfspacing]{setspace}

\usepackage{amsmath, amsfonts}
\usepackage[ruled, vlined, linesnumbered]{algorithm2e}
\usepackage{graphicx}
\usepackage{textcomp}
\usepackage{xcolor}
\usepackage{booktabs}
\usepackage{multirow}
\usepackage[super]{nth}
\usepackage{tikz}
\newcommand*\circled[1]{\tikz[baseline=(char.base)]{
		\node[shape=circle,draw,inner sep=0.2pt] (char){#1};}}
\newcommand*\circledB[1]{\tikz[baseline=(char.base)]{
            \node[shape=circle,fill,inner sep=0.2pt] (char) {\textcolor{white}{#1}};}}
\usepackage{soul}
\usepackage{setspace}

\usepackage{fancyhdr}
\fancypagestyle{firstpage}
{
  \fancyhf{}
  \vspace{-1.5cm}
  \chead{\small Accepted for publication at Frontiers in Robotics and AI - Section Robot Vision and Artificial Perception \vspace{-1.25cm}}
  \cfoot{\thepage}
}

\def\keyFont{\fontsize{8}{11}\helveticabold}
\def\firstAuthorLast{Rachmad Vidya Wicaksana Putra {et~al.}} 

\def\Authors{Rachmad Vidya Wicaksana Putra$^{*}$, Alberto Marchisio, and Muhammad Shafique}

\def\Address{eBrain Lab, Division of Engineering, New York University (NYU) Abu Dhabi, Abu Dhabi, United Arab Emirates  
}

\def\corrAuthor{Rachmad Vidya Wicaksana Putra}
\def\corrEmail{rachmad.putra@nyu.edu}

\begin{document}
\onecolumn
\firstpage{1}

\title[SNN4Agents: SNNs for Autonomous Agents]
{SNN4Agents: A Framework for Developing Energy-Efficient Embodied Spiking Neural Networks for Autonomous Agents} 

\author[\firstAuthorLast]{\Authors} 
\address{} 
\correspondance{} 


\extraAuth{}

\maketitle
\thispagestyle{firstpage}

\vspace{-0.7cm}
\begin{abstract}
Recent trends have shown that autonomous agents, such as Autonomous Ground Vehicles (AGVs), Unmanned Aerial Vehicles (UAVs), and mobile robots, effectively improve human productivity in solving diverse tasks. 
However, since these agents are typically powered by portable batteries, they require extremely low power/energy consumption to operate in a long lifespan. 
To solve this challenge, neuromorphic computing has emerged as a promising solution, where bio-inspired Spiking Neural Networks (SNNs) use spikes from event-based cameras or data conversion pre-processing to perform sparse computations efficiently. 
However, the studies of SNN deployments for autonomous agents are still at an early stage. 
Hence, the optimization stages for enabling efficient embodied SNN deployments for autonomous agents have not been defined systematically.
Toward this, we propose a novel framework called \textbf{SNN4Agents} that consists of a set of optimization techniques for designing energy-efficient embodied SNNs targeting autonomous agent applications. 
Our SNN4Agents employs weight quantization, timestep reduction, and attention window reduction to jointly improve the energy efficiency, reduce the memory footprint, optimize the processing latency, while maintaining high accuracy. 
In the evaluation, we investigate use cases of event-based car recognition, and explore the trade-offs among accuracy, latency, memory, and energy consumption. 
The experimental results show that our proposed framework can maintain high accuracy (i.e., 84.12\% accuracy) with 68.75\% memory saving, 3.58x speed-up, and 4.03x energy efficiency improvement as compared to the state-of-the-art work for the NCARS dataset. 
In this manner, our SNN4Agents framework paves the way toward enabling energy-efficient embodied SNN deployments for autonomous agents. 

\tiny
\keyFont{\section{Keywords:} Neuromorphic computing, spiking neural networks, autonomous agents, automotive data, neuromorphic processor, energy efficiency.} 
\end{abstract}

\section{Introduction}
\label{Sec_Intro}

In recent years, the interest in implementing neuromorphic artificial intelligence based on Spiking Neural Networks (SNNs) for autonomous agents (so-called \textit{SNN-based autonomous agents}) has rapidly increased~\citep{Ref_Putra_EmbodiedNeuroAI4Robotics_arXiv24, Ref_Bartolozzi_EmbodiedNeuroIntel_Nature22}. 
The reason is that, SNNs can offer high accuracy due to effective learning mechanism~\citep{Ref_Rathi_SNNsurvey_CSUR23, Ref_Putra_SpikeDyn_DAC21, Ref_Putra_SpikeNAS_arXiv24}, low computation latency due to efficient neural/spike coding~\citep{Ref_Guo_NeuralCoding_FNINS21}, and ultra low power/energy consumption due to sparse spike-based operations~\citep{Ref_Putra_FSpiNN_TCAD20, Ref_Schuman_OpportunityNeuro_Nature22}.
To realize such systems in real life, capabilities of solving machine learning (ML) tasks like image classification~\citep{Ref_Putra_SparkXD_DAC21, Ref_Putra_EnforceSNN_FNINS22, Ref_Putra_RescueSNN_FNINS23}, object detection~\citep{Ref_Viale_CarSNN_IJCNN21, Ref_Cordone_ObjDetSNN_IJCNN22}, or object segmentation~\citep{Ref_Li_SpiCalib_arXiv22} from images/videos are required. 
Besides such functionalities, SNN-based autonomous agents also require (1) small memory footprint as they typically employ resource-constrained hardware platforms, (2) low power/energy consumption to preserve the battery lifespan as they are typically powered by portable batteries, and (3) real-time output with high accuracy to provide quick decision~\citep{Ref_Bonnevie_DynamicEnv_ICARA21, Ref_Putra_lpSpikeCon_IJCNN22, Ref_Putra_TopSpark_IROS23}.
To maximize the benefits of SNN sparse operations, event-based data can be employed as it directly provides a compatible data format for SNN processing and minimizes the pre-processing stage, such as the data-to-spike conversion (e.g., pixel data to spike train) and the spike coding. 
Therefore, the developments of SNN-based autonomous agents also need to consider event-based data, such as the NCARS dataset~\citep{Ref_Sironi_HATS_CVPR18, Ref_Bano_SNNParameters_arXiv24}.

Motivated by the above-mentioned potentials of SNN-based autonomous agents, \textit{\textbf{the targeted research problem} is how to systematically develop energy-efficient SNN-based autonomous agents considering event-based data workload.} 
An efficient solution to this problem will enable SNN-based autonomous agents to achieve high accuracy with small memory footprint, low processing latency, and low energy consumption.

\subsection{State-of-the-Art Works and Their Limitations}
\label{Sec_Intro_SOTA}

The study for developing SNN-based autonomous agents is still at an early stage. 
The state-of-the-art works are summarized in Table~\ref{Tab_SOTA}.
Here, we observe that most of the existing works focus on proposing frameworks and/or techniques for achieving high accuracy~\citep{Ref_Putra_lpSpikeCon_IJCNN22, Ref_Putra_Mantis_ICARA23, Ref_Putra_TopSpark_IROS23}. 
However, these works have not considered event-based data workloads, therefore requiring a relatively complex pre-processing stage, including data-to-spike conversion and spike coding. 
Some other works explore techniques for achieving high accuracy considering event-based automotive data~\citep{Ref_Viale_CarSNN_IJCNN21, Ref_Viale_LaneSNN_IROS22, Ref_Cordone_ObjDetSNN_IJCNN22}. 
However, these works have not considered optimizing the model size to fit into the resource-constrained autonomous agents.  
All the above-discussed limitations of the state-of-the-art expose that, \textit{the optimization stages for enabling efficient SNN deployments for autonomous agents have not been defined systematically}.

\begin{table}[h]
\caption{State-of-the-art for SNN-based autonomous agents}
\vspace{0.2cm}
\small
\begin{tabular}{|c|l|}
\hline
\textbf{Work} & \multicolumn{1}{c|}{\textbf{Key Attributes}} \\ \hline
\begin{tabular}[c]{@{}c@{}}Mantis\\ \citep{Ref_Putra_Mantis_ICARA23} \end{tabular} & \begin{tabular}[c]{@{}l@{}}+ It presents a set of techniques for optimizing SNNs for autonomous agents.\\ - It considers conventional non-event data, thus its efficiency is sub-optimal.\end{tabular} \\ \hline
\begin{tabular}[c]{@{}c@{}}lpSpikeCon\\ \citep{Ref_Putra_lpSpikeCon_IJCNN22} \end{tabular} & \begin{tabular}[c]{@{}l@{}}+ It enables low-precision SNNs under unsupervised continual learning settings.\\ - It considers conventional non-event data, thus its efficiency is sub-optimal.\end{tabular} \\ \hline
\begin{tabular}[c]{@{}c@{}}TopSpark\\ \citep{Ref_Putra_TopSpark_IROS23} \end{tabular}  & \begin{tabular}[c]{@{}l@{}}+ It optimizes the computation time for both training and inference.\\ - It considers conventional non-event data, thus its efficiency is sub-optimal.\end{tabular} \\ \hline
\begin{tabular}[c]{@{}c@{}}CarSNN\\ \citep{Ref_Viale_CarSNN_IJCNN21} \end{tabular} & \begin{tabular}[c]{@{}l@{}}+ It employs the STBP~\citep{Ref_Wu_STBP_FNINS18} for learning event-based car data.\\ - It does not consider optimization for resource-constrained autonomous agents.\end{tabular} \\ \hline
\begin{tabular}[c]{@{}c@{}}LaneSNN\\ \citep{Ref_Viale_LaneSNN_IROS22} \end{tabular}  & \begin{tabular}[c]{@{}l@{}}+ It employs the STBP~\citep{Ref_Wu_STBP_FNINS18} for learning event-based road lane data.\\ - It does not consider optimization for resource-constrained autonomous agents.\end{tabular} \\ \hline
\begin{tabular}[c]{@{}c@{}}Object Detection SNN\\ \citep{Ref_Cordone_ObjDetSNN_IJCNN22} \end{tabular} & \begin{tabular}[c]{@{}l@{}}+ It employs voxel cube and surrogate gradient for learning event-based data.\\ - It does not consider optimization for resource-constrained autonomous agents.\end{tabular}   \\ \hline
\begin{tabular}[c]{@{}c@{}}\textbf{\textit{SNN4Agents}}\\ \textbf{\textit{(ours)}}\end{tabular}     & \begin{tabular}[c]{@{}l@{}}+ It presents a set of techniques for optimizing SNNs for autonomous agents. \\ + It considers event-based data to enable efficient input activation.\end{tabular}             \\ \hline
\end{tabular}
\label{Tab_SOTA}
\end{table}

\subsection{Motivational Case Study}
\label{Sec_Intro_CaseStudy}

\begin{figure}[t]
\centering
\includegraphics[width=\linewidth]{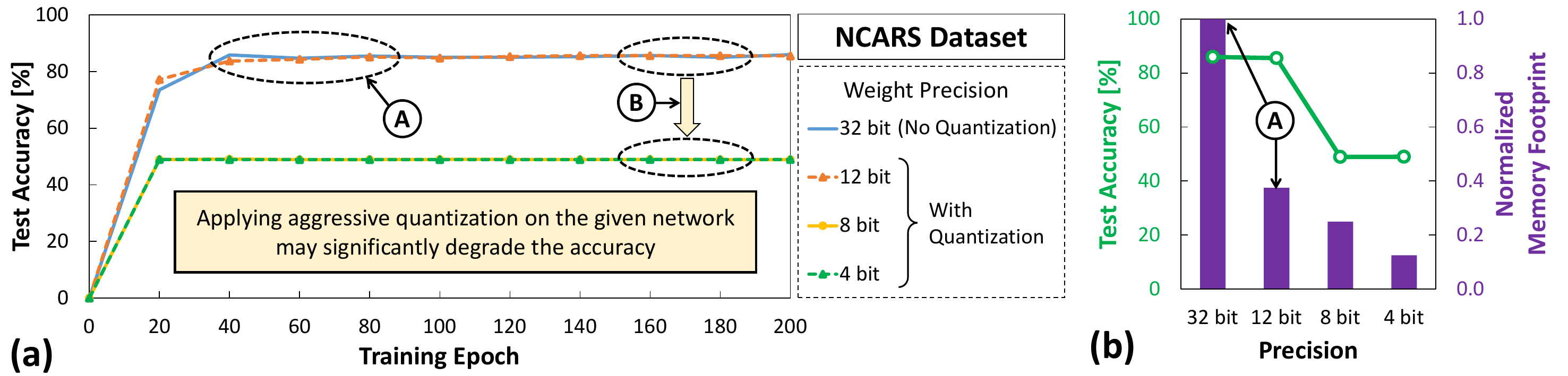}
\caption{Experimental results for observing the impact of different weight precision levels (i.e., 32, 12, 8, and 4 bits) on: \textbf{(a)}  the accuracy of an SNN model across the training epochs; and \textbf{(b)} the accuracy after 200 training epoch and the corresponding memory footprints.} 
\label{Fig_ObserveQuant}
\end{figure}

To highlight the potentials of further optimizing SNNs for autonomous agents from the current state-of-the-art, we perform an experimental case study that considers applying different levels of weight quantization (i.e., precision) to an SNN model from the work of~\cite{Ref_Viale_CarSNN_IJCNN21} with event-based NCARS dataset. 
Details of the experimental setup will be discussed further in Section~\ref{Sec_Eval}. 
The experimental results shown in Figure~\ref{Fig_ObserveQuant} show several key observations as discussed in the following.
\begin{itemize}
    \item[\circled{A}] Quantization with 12-bit precision level achieves comparable accuracy to the original 32-bit precision level (without quantization) across training epochs, while offering 2.7x memory saving.
    \item [\circled{B}] Quantization settings with 8-bit and 4-bit precision levels suffer from significant accuracy degradation as they roughly have 50\% accuracy across training epochs. 
    Considering that the NCARS dataset has 2 classes (i.e., `car' or `background'), these results indicate that the network performs random guessing as it is not properly trained.  
\end{itemize}

\vspace{-0.4cm}

These observations show that there is an opportunity to optimize further the SNN models to make them fit into the resource-constrained autonomous agents. 
However, simply performing aggressive quantization on the given network may significantly degrade the accuracy. 
Therefore, \textit{\textbf{the research challenge} is how to effectively perform different optimization techniques without significantly degrading the accuracy}.

\subsection{Our Novel Contributions}
\label{Sec_Intro_Novelty}

\begin{figure}[hbtp]
\centering
\includegraphics[width=0.78\linewidth]{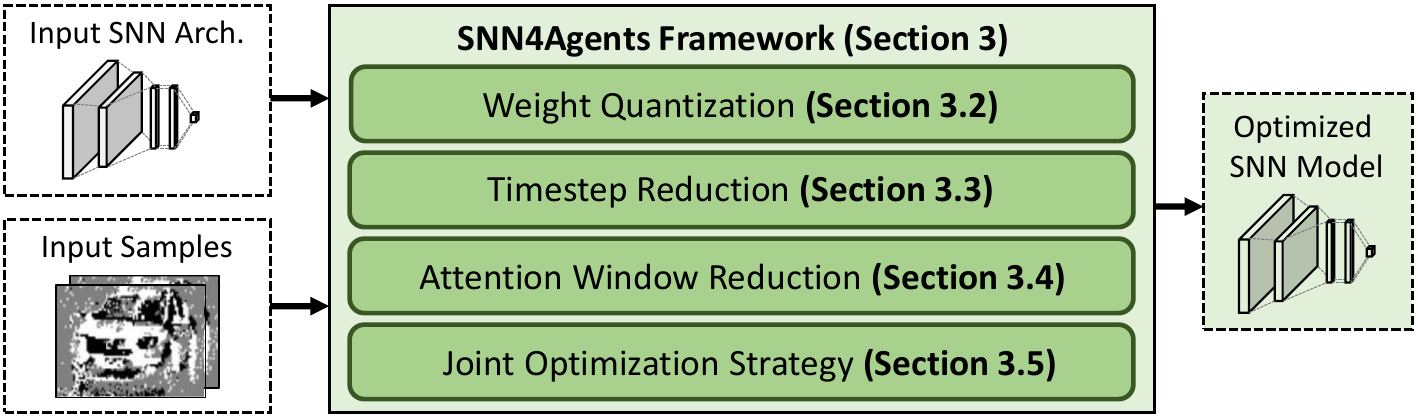}
\caption{The overview of our novel contributions, shown in green boxes} 
\label{Fig_Novelty}
\end{figure}

To address the targeted research problem and the related challenges, we propose \textit{a novel framework called \textbf{SNN4Agents} for developing energy-efficient embodied SNNs for autonomous agents}.
The overview of our SNN4Agents framework is shown in Figure~\ref{Fig_Novelty} and its key steps are briefly described in the following.
\begin{itemize}
    \item \textbf{Weight Quantization (Section~\ref{Sec_Frame_Quantize}):} 
    It aims to find the appropriate quantization settings that meet the memory constraint, we perform design space exploration for different precision levels while observing their impact on the accuracy.  
    \item \textbf{Timestep Reduction (Section~\ref{Sec_Frame_Timestep}):} 
    It aims to find the appropriate processing timesteps that meet the latency constraint, we perform design space exploration for different timestep values while observing their impact on the accuracy. 
    \item \textbf{Attention Window Reduction (Section~\ref{Sec_Frame_AttWindow}):} 
    It aims to reduce the compute requirements and hence the processing power/energy, we explore different attention window sizes from the input samples and observe their impact on the accuracy. 
    \item \textbf{Joint Optimization Strategy (Section~\ref{Sec_Frame_Joint}):} 
    It aims to maximize the benefits from the previous individual optimization steps, we perform a joint optimization strategy by leveraging design space exploration that considers the appropriate settings from individual optimization steps.  
\end{itemize}

\vspace{-0.2cm}

\section{Preliminaries}
\label{Sec_Prelim}

\subsection{Spiking Neural Networks (SNNs)}
\label{Sec_Prelim_SNNs}

\subsubsection{Overview}

\begin{figure}[hbtp]
\vspace{-0.2cm}
\centering
\includegraphics[width=\linewidth]{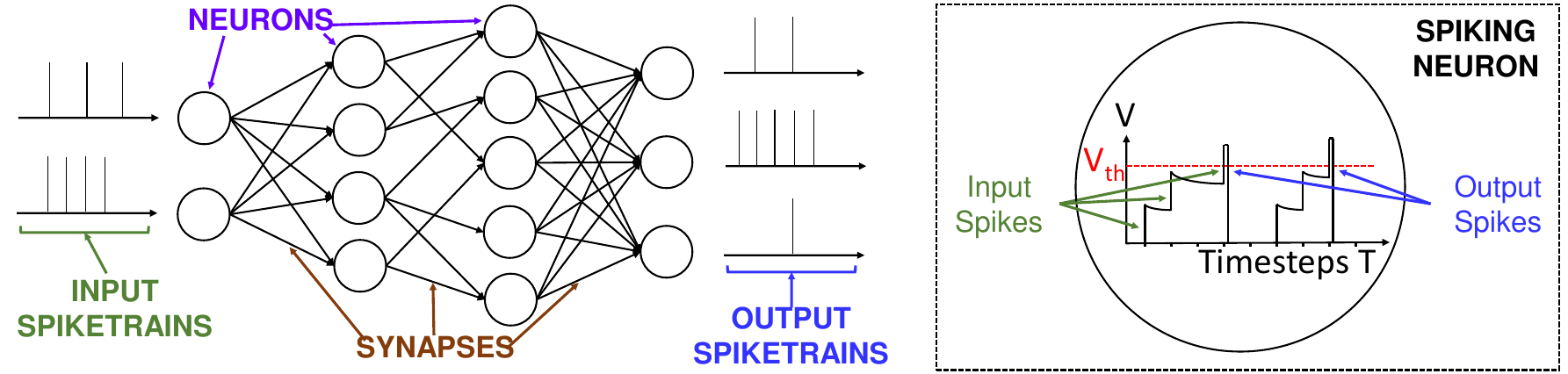}
\vspace{-0.7cm}
\caption{Overview of the functionality of a Spiking Neural Network.} 
\label{Fig_SNN}
\vspace{-0.3cm}
\end{figure}

Spiking Neural Networks (SNNs) are considered the third generation neural networks~\citep{Ref_Maass_SNN_NeuNet97}. 
An overview of the SNNs' functionality is shown in Figure~\ref{Fig_SNN}. 
The input spike trains are processed by the spiking neurons and the information is propagated to the neurons in the following layers. 
Among the most popular spiking neuron models, the Leaky-Integrate-and-Fire (LIF)~\citep{Ref_Izhikevich_CompareModels_TNN04, Wang_2014EMBC_LeakyIntegrateAndFire} emerges as an efficient trade-off between complexity and plausibility. 
The operational time of a neuron to process a spike train from a single input data (e.g., an image pixel) is defined as \textit{timestep}~\citep{Ref_Putra_TopSpark_IROS23}. 
At each timestep, the membrane potential $V_m$ of the neuron increases for each incoming input spike. 
When $V_m$ overcomes the voltage threshold $V_{th}$, the neuron emits an output spike~\citep{Ref_Roy_SpikeMachineIntel_Nature19, Ref_Putra_ReSpawn_ICCAD21, Ref_Putra_SoftSNN_DAC22}. 
In this way, the spikes propagate from input to output and multiple layers of spiking neurons in a network.

While SNNs follow the wave of success of traditional (non-spiking) Deep Neural Networks (DNNs), they offer the following advantages.  

\begin{itemize}
    \item \textit{Biological Plausibility:} 
    The SNNs' functionality is inspired by the behavior of the biological brain, where spikes are propagated across neurons for conveying information. 
    This may open possibilities of cognition and robustness for solving diverse machine learning tasks.
    \item \textit{Ultra-Low Power/Energy Consumption:} 
    The dynamic power in SNNs is consumed only in the presence of spikes, hence offering ultra low processing power/energy in a long time operational period. 
    \item \textit{Efficient Interface with Event-based Sensors:} 
    The event sequences captured by event-based sensors can directly be utilized as the input of SNNs without complex pre-processing (e.g., data-to-spike conversion). 
\end{itemize}

Besides these advantages, it is actually challenging to efficiently train SNNs due to the non-differentiability of the spiking loss function~\citep{Ruckauer_2019arxiv_NonDifferentiableLossFunctionSNN}. 
Hence, to overcome this challenge, two possible techniques have been proposed in the literature. 
(1) The \textit{DNN-to-SNN conversion} approach~\citep{Ref_Hao_ANNtoSNNCalibration_ICLR23, Ref_Bu_ANNtoSNNConversion_ICLR22} trains a DNN and then converts the model into the equivalent spiking counterpart. 
(2) The \textit{direct SNN training} approach~\citep{Ref_Neftci_SurrogateSNNs_IEEEMSP19} employs a surrogate gradient function to approximate the spiking loss function, in such a way that it can be differentiated and incorporated into the backpropagation flow. 
Since approach-(1) requires DNN training, it cannot be directly used when dealing with event-based data~\citep{Massa_2020IJCNN_EfficientSNNGestures}. 
Moreover, it typically requires a larger number of timesteps than approach-(2)~\citep{Ref_Chowdhury_OneTimestepIsAllYouNeed_arxiv21}.
Therefore, \textit{in this work, we consider employing direct SNN training, i.e., Spatio-Temporal Back-Propagation (STBP) learning rule, which leverages both spatial and temporal information within the streaming spikes}~\citep{Ref_Wu_STBP_FNINS18, Ref_Viale_CarSNN_IJCNN21}.

\subsection{Quantization}
\label{Sec_Prelim_Quantize}

Quantization is a prominent optimization technique which can effectively compress SNN models with relatively low overhead, since it only needs to reduce the data precision~\citep{Ref_Gupta_Precision_ICML15, Ref_Micikevicius_Precision_ICLR18}. 
Implementation of quantization requires a specific setting that encompasses \textit{quantization scheme} and \textit{rounding scheme} as discussed in the following. 

\begin{figure}[hbtp]
\vspace{-0.2cm}
\centering
\includegraphics[width=\linewidth]{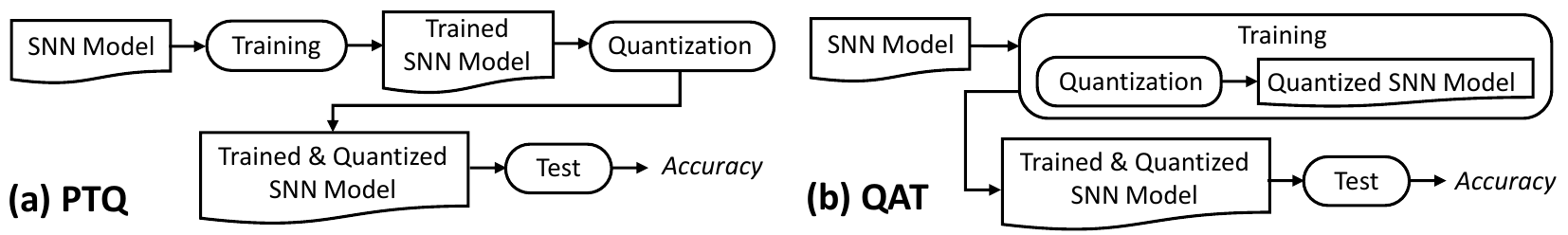}
\vspace{-0.7cm}
\caption{The flow of (a) PTQ: Post-Training Quantization, and (b) QAT: Quantization-Aware Training.} 
\label{Fig_Quantize}
\vspace{-0.3cm}
\end{figure}

\textbf{Quantization Schemes:}
There are two widely used quantization schemes for SNNs: \textit{Post-Training Quantization (PTQ)}, and \textit{Quantization-Aware Training (QAT)}~\citep{Ref_Putra_QSpiNN_IJCNN21}; see the illustration of PTQ and QAT flows in Figure~\ref{Fig_Quantize}. 
PTQ typically trains the given network with a floating-point precision, such as 32-bit floating point (FP32), and then the quantization is applied to the trained SNN model with the given precision level, resulting in a quantized SNN model for the inference phase. 
Meanwhile, QAT typically performs quantization to the given network with the given precision level during the training phase, resulting in a trained and quantized SNN model which can be directly used for the inference phase. 
PTQ process is typically simpler and more efficient than QAT as it performs quantization once after the training phase is finished. 

\begin{figure}[t]
\centering
\includegraphics[width=\linewidth]{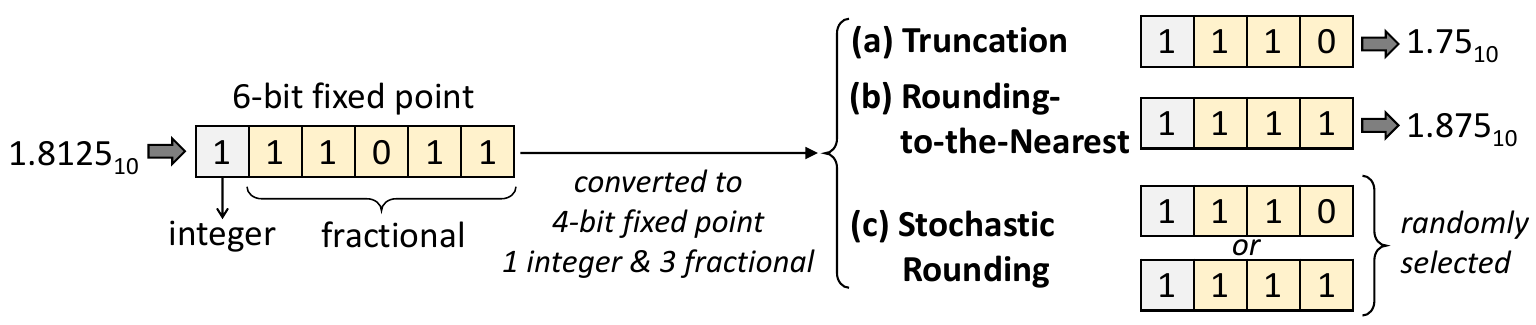}
\vspace{-0.7cm}
\caption{Illustration of different rounding schemes: \textit{Truncation (TR)}, \textit{Rounding-to-the-Nearest (RN)}, and \textit{Stochastic Rounding (SR)}; based on studies in~\citep{Ref_Putra_QSpiNN_IJCNN21}.} 
\label{Fig_Rounding}
\vspace{-0.3cm}
\end{figure}

\textbf{Rounding Schemes:}
The implementation of quantization typically requires a specific rounding scheme for determining how the value will be curtailed. 
From the literature, there are three widely used rounding schemes for SNN models: \textit{Truncation (TR)}, \textit{Rounding-to-the-Nearest (RN)}, and \textit{Stochastic Rounding (SR)}~\citep{Ref_Putra_QSpiNN_IJCNN21}. 
Illustration of these rounding schemes is shown in Figure~\ref{Fig_Rounding}
Among these rounding schemes, TR has the simplest operation since it simply keeps the defined number of the most significant bits and discards the other remaining bits.  

In this work, \textit{we employ PTQ scheme with TR rounding since their combination can quickly provide representative results for different quantization settings, thereby enabling fast design space exploration (DSE), which is important for our studies}. 

\subsection{Event-Based Automotive Data}
\label{Sec_Prelim_AutoData}

\textbf{Prophesee NCARS Dataset~\citep{Ref_Sironi_HATS_CVPR18}:} 
This event-based dataset contains a collection of 24K samples that have a duration of 100 ms each, recorded with the Asynchronous Time-based Image Sensor (ATIS) camera~\citep{Posch_2011JSSC_ATISSensor}. Each sample, labeled as either ``car'' or ``background'', is encoded as a sequence of events that contains the following information:

\begin{itemize}
    \item the timestamp $t$ of when the event occurred;
    \item the spatial coordinates $x$ and $y$ of the pixel;    
    \item the polarity $p$ of the brightness variation, which can either be positive or negative.
\end{itemize}

The data is split into 15,422 training and 8,607 testing samples. Each sample has variable sizes and can be cropped. Based on the distribution of events, we can identify an \textit{attention window}, i.e., a region in which the events are more concentrated. Typical sizes of the attention windows used in state-of-the-art works~\citep{Ref_Viale_CarSNN_IJCNN21} can scale down to 100$\times$100 or 50$\times$50 and significantly reduce the computational and memory requirements of the SNN, without compromising the accuracy much.

\subsection{SNN Architecture}

SNN architectures are composed of a sequence of layers of spiking neurons. Their structure and connections define the type of layers. 
In this work, we employ the CarSNN architectures~\citep{Ref_Viale_CarSNN_IJCNN21} that efficiently execute car recognition with STBP-based learning rule. 
We implement two SNN architectures built with different sizes of attention window. Both models are composed of an interleaved sequence of two convolutional layers and three average pooling layers, followed by two fully-connected layers. The first SNN architecture, described in Table~\ref{Tab_SNNarch_100x100}, has a 100$\times$100 attention window, while the second SNN architecture, described in Table~\ref{Tab_SNNarch_50x50}, has a 50$\times$50 attention window. Note that different attention window sizes affect the feature map sizes of each SNN layer. Hence, these architectures have different numbers of input and output channels in the fully-connected layers than the latter.


\vspace{-0.5cm}
\begin{table}[hbtp]
\caption{SNN architecture with 100$\times$100 attention window, based on the work of~\cite{Ref_Viale_CarSNN_IJCNN21}.}
\vspace{0.2cm}
\label{Tab_SNNarch_100x100}
\small
\begin{tabular}{|c|c|c|c|c|c|}
\hline
\textbf{\begin{tabular}[c]{@{}c@{}}Layer Type\end{tabular}} & \textbf{\begin{tabular}[c]{@{}c@{}}Input Channel\end{tabular}} & \textbf{\begin{tabular}[c]{@{}c@{}}Output Channel\end{tabular}} & \textbf{\begin{tabular}[c]{@{}c@{}}Kernel Size\end{tabular}} & \textbf{Padding} & \textbf{Stride} \\ \hline 
Average Pooling & 2 & 2 & 4 & - & - \\ \hline
Convolution & 2 & 32 & 3 & 1 & 1 \\ \hline
Average Pooling & 32 & 32 & 2 & - & - \\ \hline
Convolution & 32 & 32 & 3 & 1 & 1 \\ \hline
Average Pooling & 32 & 32 & 2 & - & - \\ \hline
Fully-Connected & 1152 & 512 & - & - & - \\ \hline
Fully-Connected & 512 & 2 & - & - & - \\ \hline
\end{tabular}
\end{table}

\vspace{-0.5cm}
\begin{table}[hbtp]
\caption{SNN architecture with 50$\times$50 attention window, based on the work of~\cite{Ref_Viale_CarSNN_IJCNN21}.}
\vspace{0.2cm}
\label{Tab_SNNarch_50x50}
\small
\begin{tabular}{|c|c|c|c|c|c|}
\hline
\textbf{\begin{tabular}[c]{@{}c@{}}Layer Type\end{tabular}} & \textbf{\begin{tabular}[c]{@{}c@{}}Input Channel\end{tabular}} & \textbf{\begin{tabular}[c]{@{}c@{}}Output Channel\end{tabular}} & \textbf{\begin{tabular}[c]{@{}c@{}}Kernel Size\end{tabular}} & \textbf{Padding} & \textbf{Stride} \\ \hline 
Average Pooling & 2 & 2 & 4 & - & - \\ \hline
Convolution & 2 & 32 & 3 & 1 & 1 \\ \hline
Average Pooling & 32 & 32 & 2 & - & - \\ \hline
Convolution & 32 & 32 & 3 & 1 & 1 \\ \hline
Average Pooling & 32 & 32 & 2 & - & - \\ \hline
Fully-Connected & 288 & 144 & - & - & - \\ \hline
Fully-Connected & 144 & 2 & - & - & - \\ \hline
\end{tabular}
\end{table}

\section{Our SNN4Agents Framework}
\label{Sec_Frame}

\subsection{Overview}
\label{Sec_Frame_Overview}

\begin{figure}[hbtp]
\centering
\includegraphics[width=\linewidth]{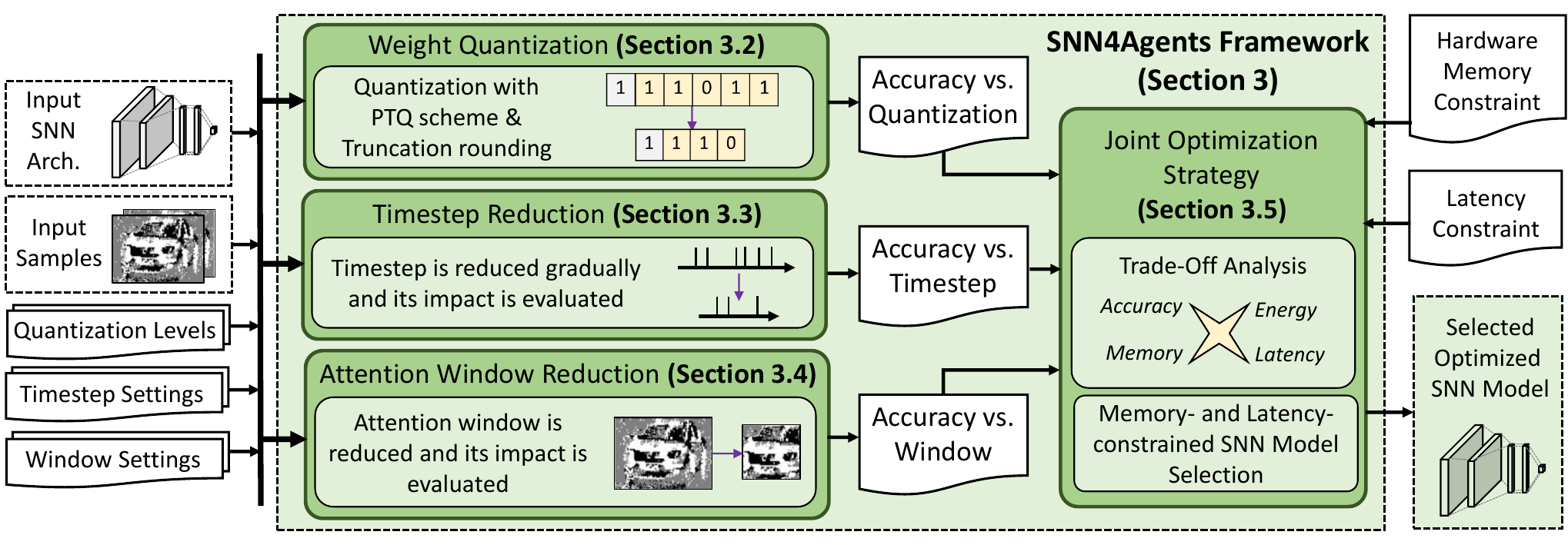}
\caption{The flow of our SNN4Agents framework, with the technical contributions highlighted in green.} 
\label{Fig_SNN4Agents}
\end{figure}

Our SNN4Agents framework employs a set of optimization techniques targeting different design aspects, including \textit{model compression through weight quantization}, \textit{latency optimization through timestep reduction}, \textit{input data optimization through attention window reduction}; and then performs \textit{joint optimization strategy} to maximize benefits from the individual optimization techniques.
The flow of our SNN4Agents framework is illustrated in Figure~\ref{Fig_SNN4Agents}, and the detailed discussion for its steps are provided in Section~\ref{Sec_Frame_Quantize} - Section~\ref{Sec_Frame_Joint}.

\subsection{Model Compression through Quantization}
\label{Sec_Frame_Quantize}

\vspace{-0.2cm}
To effectively compress the model size, we perform weight quantization through PTQ with TR rounding scheme. 
To do this, we first train the given network without quantization, while employing baseline settings for timestep, attention window, and training epoch. 
For this scenario, we employ 32-bit precision, 20 timestep, 100$\times$100 attention window, and 200 training epoch. 
Once the training phase is finished, we perform quantization process to the trained network. 
Afterward, we perform DSE under different weight precision levels (i.e., 32, 16, 12, 10, 8, 6, and 4 bit) to evaluate their impact on the accuracy; see the parameter settings for DSE in Table~\ref{Tab_Setting_Quantize}. 

\vspace{-0.3cm}
\begin{table}[hbtp]
\caption{Parameter settings for exploring the impact of different precision levels.}
\vspace{0.2cm}
\label{Tab_Setting_Quantize}
\small
\begin{tabular}{|c|c|c|c|}
\hline
\textbf{\textit{Precision (bit)}} & \textbf{Timestep} & \textbf{Attention Window} & \textbf{Training Epoch} \\ \hline
32, 16, 12, 10, 8, 6, and 4 & 20 & 100$\times$100 & 0-200 \\ \hline
\end{tabular}
\end{table}

\begin{figure}[hbtp]
\vspace{-0.2cm}
\centering
\includegraphics[width=0.8\linewidth]{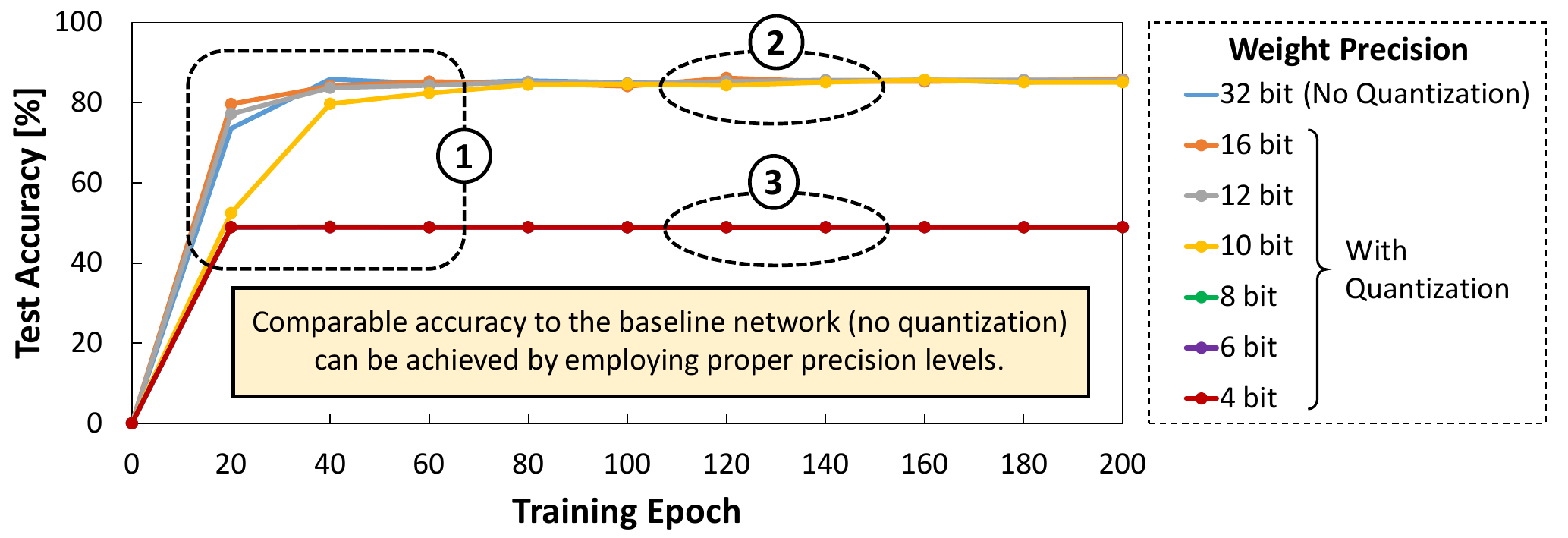}
\caption{Results of accuracy across different precision levels (i.e., 32, 16, 12, 10, 8, and 4 bit).} 
\label{Fig_QuantizeExp}
\vspace{-0.2cm}
\end{figure}

Experimental results of DSE are shown in Figure~\ref{Fig_QuantizeExp}, from which we draw the following key observations.
\begin{itemize}
    \item In the early of training phase (e.g., $\leq$ 60 training epoch), the network is still learning new information, hence the accuracy curve is increasing for 16-, 12-, and 10-bit precision levels, as shown by \circled{1}.
    \item Employing 16-, 12-, and 10-bit precision levels for SNN weights lead to comparable accuracy to the original SNN model with 32-bit precision (no quantization) after running at least 80 training epoch, as shown by \circled{2}.
    \item Employing 8-, 6-, and 4-bit precision levels for SNN weights lead to significant accuracy degradation, as they can only reach about 50\% accuracy across training epochs, as shown by \circled{3}. 
    These results indicate that the network is not properly trained.
\end{itemize}

These observations expose several key design guides that we should consider when applying quantization.
First, \textit{selecting the precision level should be performed carefully}, so that it does not lead to a significant accuracy degradation which diminishes the benefits of quantization. 
Second, \textit{a 10-bit precision level offers a good trade-off between accuracy and memory footprint} as it can achieve comparable accuracy to that of the larger precision levels after running at least 80 training epoch. 

\subsection{Latency Optimization through Timestep Reduction}
\label{Sec_Frame_Timestep}

To effectively reduce the processing time, we perform timestep reduction. 
To do this, we shorten the timestep of SNN processing from the baseline settings, thereby curtailing the time window for presenting the spike trains. 
Here, we consider different timestep settings (i.e., 20, 15, 10, and 5) for exploring their impact on the accuracy. 
Once we reduce the timestep, the network is trained under baseline settings of precision level (no quantization), attention window, and training epoch. 
For this scenario, we employ 32-bit precision, 100$\times$100 attention window, and 200 training epoch. 
If we do not see notable differences from the baseline accuracy, then we may employ a smaller precision level, such as 16- and 4-bit weights, to trigger accuracy variation. 
The parameter settings for DSE are provided in Table~\ref{Tab_Setting_Timestep}. 

\vspace{-0.3cm}
\begin{table}[hbtp]
\caption{Parameter settings for exploring the impact of different timesteps.}
\vspace{0.2cm}
\label{Tab_Setting_Timestep}
\small
\begin{tabular}{|c|c|c|c|}
\hline
\textbf{Precision (bit)} & \textbf{\textit{Timestep}} & \textbf{Attention Window} & \textbf{Training Epoch}  \\ \hline
32, 16, and 4 & 20, 15, 10, and 5 & 100$\times$100 & 0-200 \\ \hline
\end{tabular}
\end{table}

\begin{figure}[hbtp]
\centering
\includegraphics[width=0.8\linewidth]{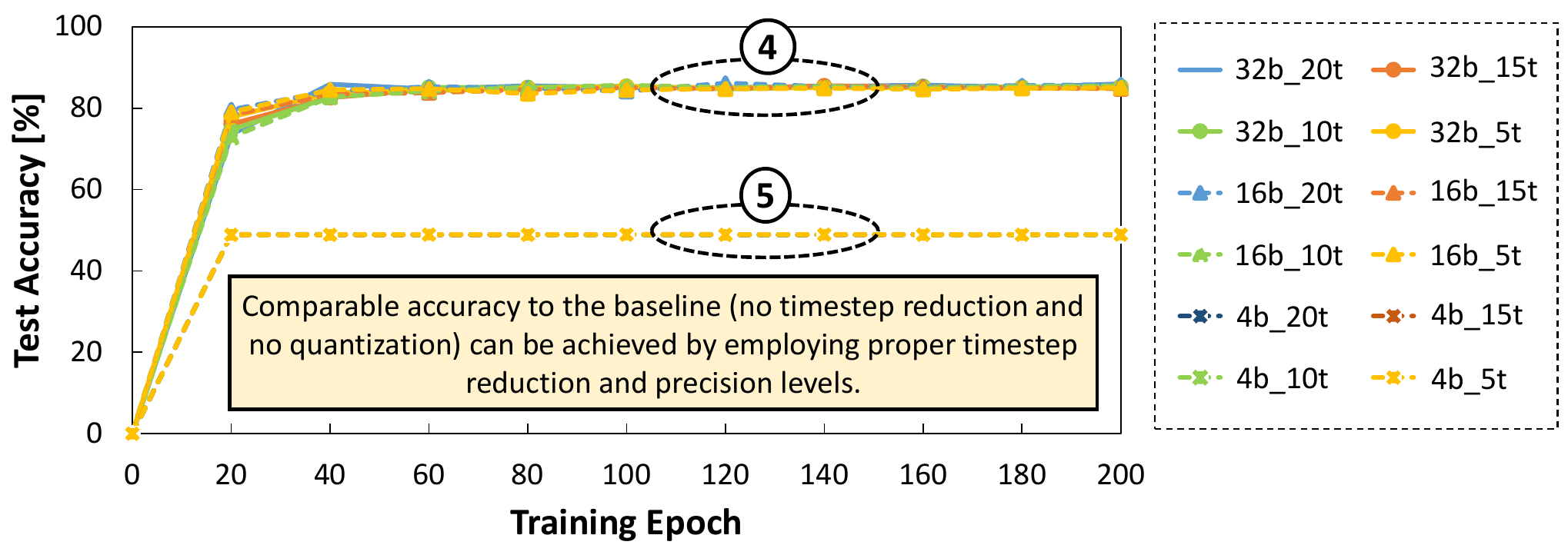}
\caption{Results of accuracy across different timestep settings (i.e., 20, 15, 10, and 5), while considering different precision levels (i.e., 32, 16, and 4 bit). Here, $B$b\_$T$t denotes $B$ bit precision and $T$ timestep.} 
\label{Fig_TimestepExp}
\end{figure}

Experimental results of DSE are shown in Figure~\ref{Fig_TimestepExp}, from which we draw the following key observations.
\begin{itemize}
    \item Employing 15, 10, and 5 timestep settings without quantization lead to comparable accuracy to the baseline model (i.e., 20 timestep without quantization) after 60 training epoch, as shown by \circled{4}. 
    Similarly, employing 20, 15, 10, and 5 timestep settings with 16-bit precision also lead to comparable accuracy to the baseline after 60 training epoch; see \circled{4}. 
    \item Employing 20, 15, 10 and 5 timestep settings with 4-bit precision level lead to significant accuracy degradation, as shown by \circled{5}. 
    It means that the 4-bit precision is relatively too small for representing temporal and spatial features from the NCARS dataset.  
\end{itemize}

\vspace{-0.3cm}

These observations show that \textit{we can apply a relatively aggressive timestep reduction (e.g., 5 timestep) without losing significant accuracy, as long as we also employ an appropriate precision level}, thereby accommodating the spatial and temporal information of the given dataset (e.g., NCARS).

\subsection{Attention Window Reduction for Input Samples}
\label{Sec_Frame_AttWindow}

We also aim at reducing the size of attention window of the input samples to optimize the computational requirements, and hence the latency and energy consumption. 
To do this, we consider different attention window sizes (i.e., 100$\times$100 and 50$\times$50), while employing baseline settings for precision level, timestep, and training epoch; see the parameter settings for DSE in Table~\ref{Tab_Setting_Window}. 

\begin{table}[hbtp]
\caption{Parameter settings for exploring impact of different attention window sizes.}
\vspace{0.2cm}
\label{Tab_Setting_Window}
\small
\begin{tabular}{|c|c|c|c|}
\hline
\textbf{Precision (bit)} & \textbf{Timestep} & \textit{\textbf{Attention Window}} & \textbf{Training Epoch} \\ \hline
32 & 20 & 100$\times$100 and 50$\times$50 & 0-200 \\ \hline
\end{tabular}
\end{table}

\begin{figure}[hbtp]
\centering
\includegraphics[width=0.6\linewidth]{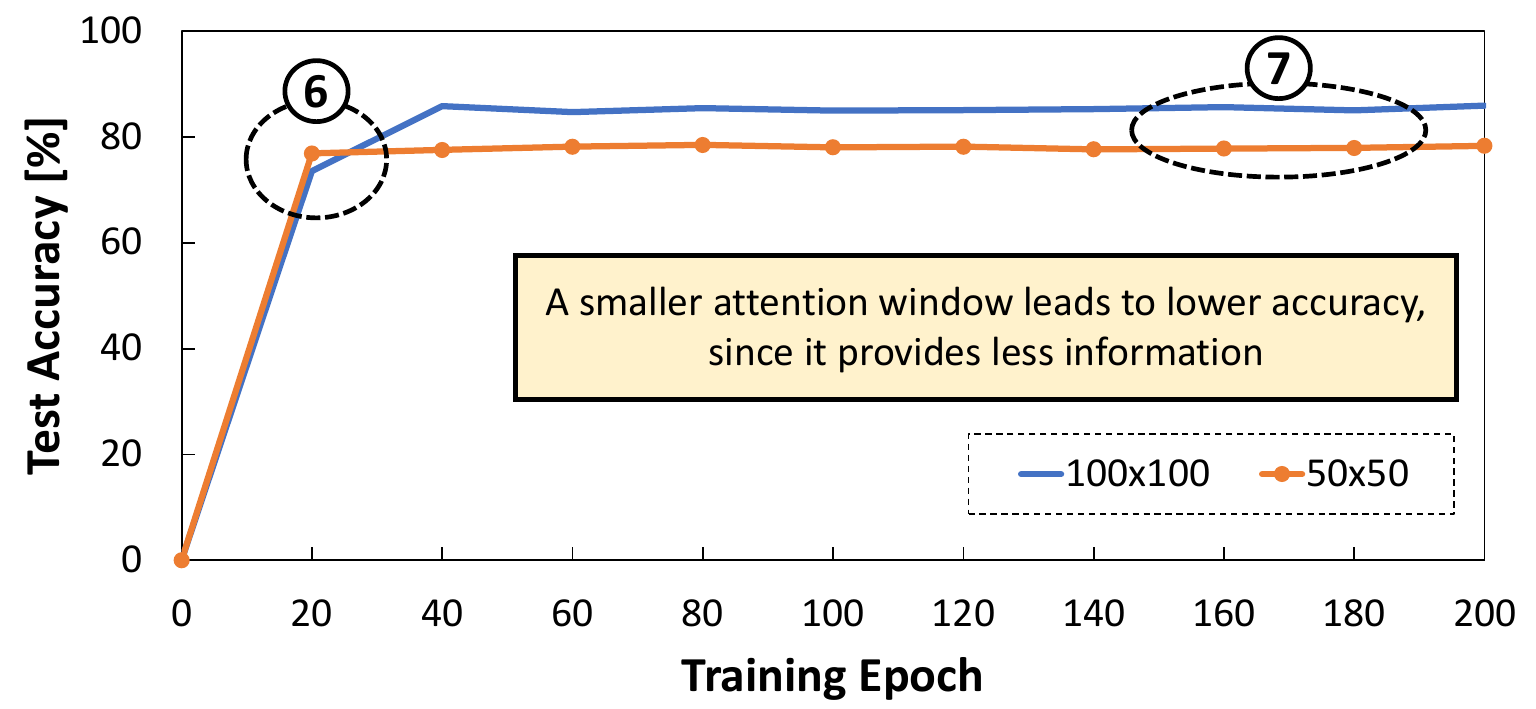}
\caption{Results of accuracy across different attention window sizes (i.e., 100$\times$100 and 50$\times$50).} 
\label{Fig_WindowExp}
\end{figure}

Experimental results of our DSE are presented in Figure~\ref{Fig_WindowExp}. 
From these results, we observe that accuracy obtained by a smaller attention window (50$\times$50) can saturate faster than a larger one (100x100), thus offering a faster training time; see \circled{6}. 
Meanwhile, a larger attention window offers better accuracy than a smaller one; see \circled{7}. 
The reason is that, a smaller attention window provides less information, thus requiring a shorter time for training the network yet limiting the accuracy that can be achieved.
These observations suggest that \textit{the reduction of attention window can be employed as long as the targeted accuracy is met}. 

\vspace{-0.3cm}

\subsection{Joint Optimization Strategy}
\label{Sec_Frame_Joint}

Each individual optimization step from previous sub-sections has demonstrated the possibility to reduce memory footprint and latency, while maintaining high accuracy. 
Therefore, \textit{to maximize these optimization benefits, we propose a strategy to jointly combine the individual optimization steps}.
Here, we leverage the key observations and design guides from previous analysis in Section~\ref{Sec_Frame_Quantize}-Section~\ref{Sec_Frame_AttWindow} to devise the following strategy.
\begin{itemize}
    \item We perform DSE for the following settings: (1) 10-32 bit precision levels of quantization, (2) 5-20 timesteps, and (3) 50$\times$50 and 100$\times$100 attention window sizes.
    \item To find the solution candidates, we analyze the experimental results for accuracy, memory footprint, latency, and energy consumption, while considering the memory and latency constraints. 
    \item If there are multiple solution candidates, we can select the most suitable one for the given constraints by trading-off the design metrics, including accuracy, memory footprint, and latency, and energy consumption. 
\end{itemize}


\section{Evaluation Methodology}
\label{Sec_Eval}

\begin{figure}[hbtp]
\centering
\includegraphics[width=0.75\linewidth]{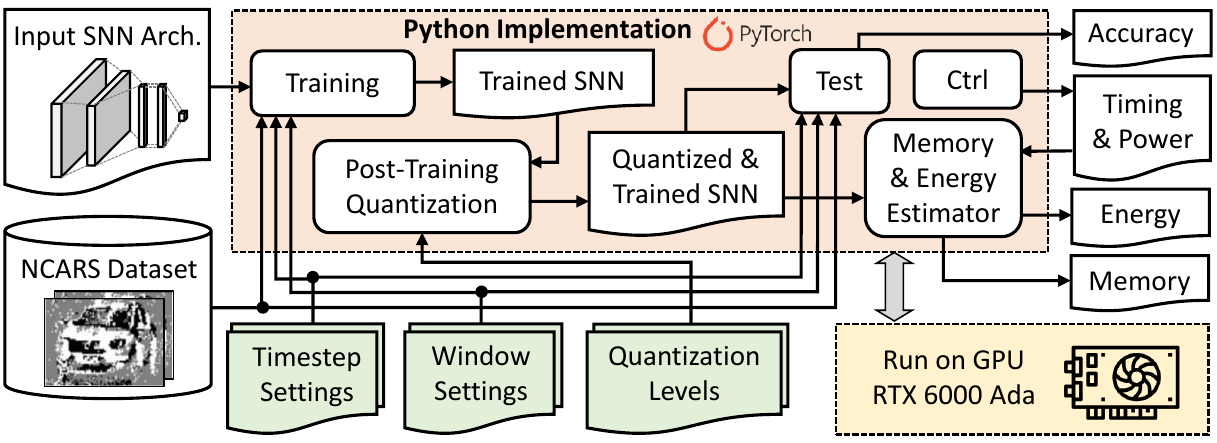}
\caption{Experimental setup for evaluating our SNN4Agents framework. The proposed settings from our SNN4Agents are incorporated into the experimental setup as highlighted in green.} 
\label{Fig_ExpSetup}
\end{figure}

To evaluate our SNN4Agents framework, we build and employ the experimental setup shown in Fig.~\ref{Fig_ExpSetup}, while considering the same evaluation conditions as widely used in the SNN community for autonomous agents~\citep{Ref_Putra_Mantis_ICARA23, Ref_Viale_CarSNN_IJCNN21}.
We use Python-based implementation and run it on Nvidia RTX 6000 Ada GPU machines for evaluating different performance metrics of our SNN4Agents, including accuracy, processing time, and power consumption. 
Then, the processing time and power consumption are leveraged to estimate the energy consumption.
Meanwhile, memory footprint is estimated by leveraging the precision level and the number of weights in the corresponding SNN architecture. 
For the case of 100$\times$100 attention window, we consider the network architecture from Table~\ref{Tab_SNNarch_100x100}; while for the case of 50$\times$50, we consider the network architecture from Table~\ref{Tab_SNNarch_50x50}. 
For the workload, we consider the NCARS dataset~\citep{Ref_Sironi_HATS_CVPR18}. 
For the DSE purpose, we consider parameter settings shown in Table~\ref{Tab_ExpSetup}, as well as the state-of-the-art work CarSNN~\citep{Ref_Viale_CarSNN_IJCNN21} as the baseline.

\vspace{-0.2cm}

\vspace{-0.3cm}
\begin{table}[hbtp]
\caption{Parameter settings for our DSE in evaluating our SNN4Agent framework.}
\vspace{0.2cm}
\label{Tab_ExpSetup}
\small
\begin{tabular}{|c|c|c|c|}
\hline
\textbf{Precision (bit)} & \textbf{Timestep} & \textbf{Attention Window} & \textbf{Training Epoch} \\ \hline
32, 16, 12, and 10 & 20, 15, 10, and 5 & 100$\times$100 and 50$\times$50 & 0-200 \\ \hline
\end{tabular}
\end{table}

Furthermore, we use several terms to represent the network model for brevity, as the following.
\begin{itemize}
    \item $B$b\_$T$t: It represents a network with $B$ bit-precision and $T$ timestep. In this case, the attention window will be explained and discussed explicitly to clearly distinguish the network model and its experimental results.
    This term will be used in Section~\ref{Sec_Res_Accuracy}, ~\ref{Sec_Res_Time}, and ~\ref{Sec_Res_Energy}.
    \item $B$b\_$W$w: It represents a network with $B$ bit-precision and $W$$\times$$W$ attention window, which encompasses different timestep settings. This term is used when discussing memory footprint in Section~\ref{Sec_Res_Mermory}, as networks with different timesteps have the same weight memory footprint.  
    \item $B$b\_$T$t\_$W$w: It represents a network with $B$ bit-precision, $T$ timestep, and $W$$\times$$W$ attention window. This term is used when discussing the trade-off analysis in Section~\ref{Sec_Res_TradeOff}.
    \item $T$t\_$W$w: It represents a network with $T$ timestep and $W$$\times$$W$ attention window. This term is used when discussing the computational complexity in Section~\ref{Sec_Res_Complexity}.
\end{itemize}

\section{Experimental Results and Discussion}
\label{Sec_Res}

\subsection{Maintaining High Accuracy}
\label{Sec_Res_Accuracy}

\begin{figure}[hbtp]
\centering
\includegraphics[width=0.9\linewidth]{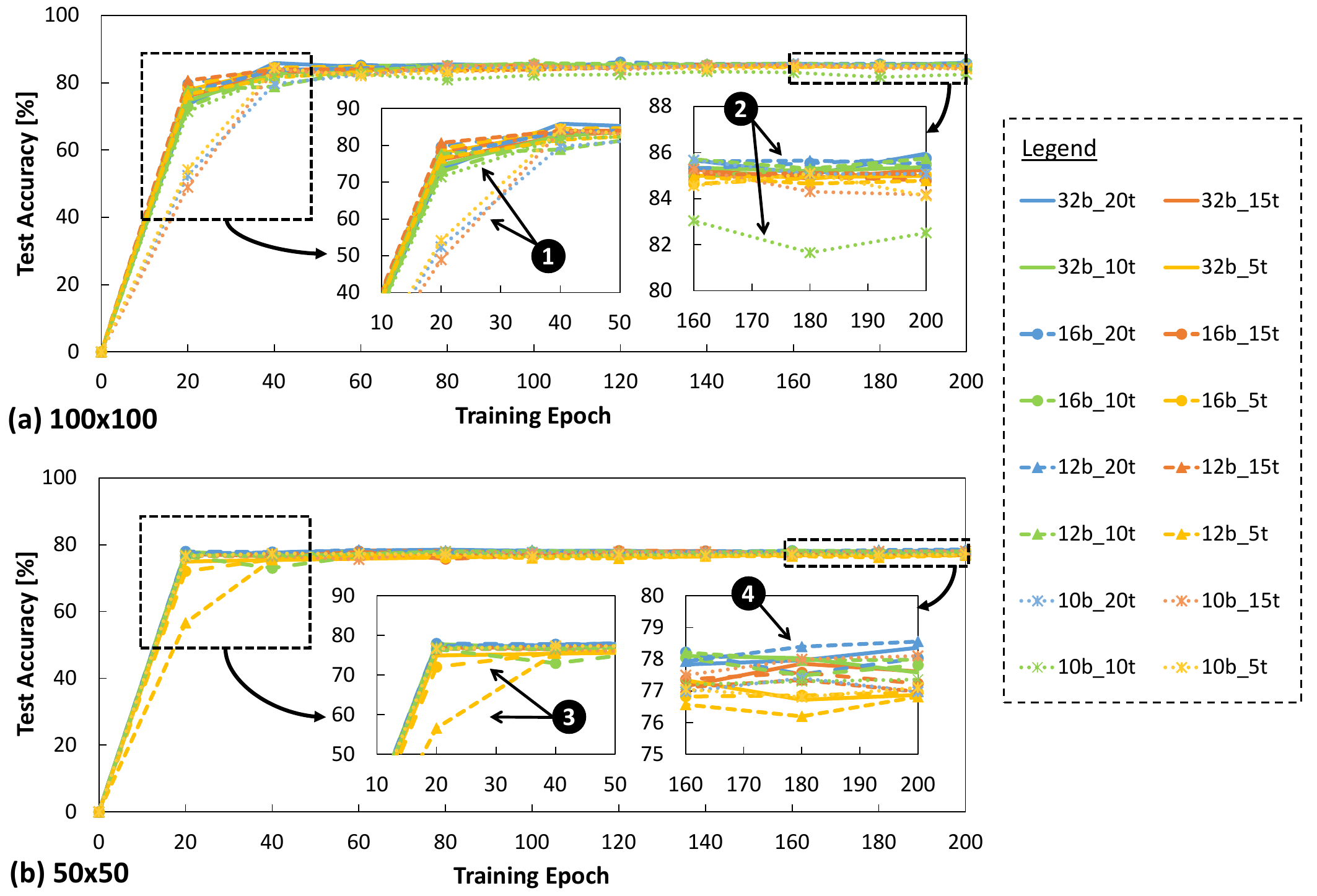}
\vspace{-0.2cm}
\caption{Experimental results for accuracy across different parameter settings, including precision levels, timesteps, and training epochs under (a) 100$\times$100 attention window, and (b) 50$\times$50 attention window.} 
\label{Fig_Result_Accuracy}
\end{figure}

Experimental results for accuracy are presented in Figure~\ref{Fig_Result_Accuracy}. 
These results show that the baseline model (32b\_20t) can achieve 85.95\% accuracy, while our optimized SNN models with 100$\times$100 attention window achieve 84.12\% - 85.76\% accuracy and our optimized SNN models with 50$\times$50 attention window achieve 76.81\% - 78.56\% accuracy.  
In the 100$\times$100 attention window case, we observe that 10-bit precision levels generally lead to lower learning quality (i.e., accuracy) across all training epochs as compared to other precision levels; see \circledB{1} and \circledB{2}.  
These indicate that smaller precision levels have smaller value representation capabilities than the larger ones, and hence they require a longer training period to achieve comparable accuracy. 
For instance, 10-bit precision levels can achieve close to 80\% accuracy in 40 training epoch, while other precision levels can achieve it in 20 training epoch.  
Due to the same reason, similar patterns are observed in the 50$\times$50 attention window case; see \circledB{3} and \circledB{4}. 
However, in this case, the accuracy scores from different precision levels are not much different, as they lay within 2\% accuracy range; see \circledB{4}.
The reason is that, a smaller attention window means less information to be conveyed and processed by different precision levels, thereby limiting the final accuracy.  

\vspace{-0.4cm}
\vspace{-0.4cm}

\subsection{Memory Savings}
\label{Sec_Res_Mermory}

\begin{figure}[h]
\centering
\includegraphics[width=0.8\linewidth]{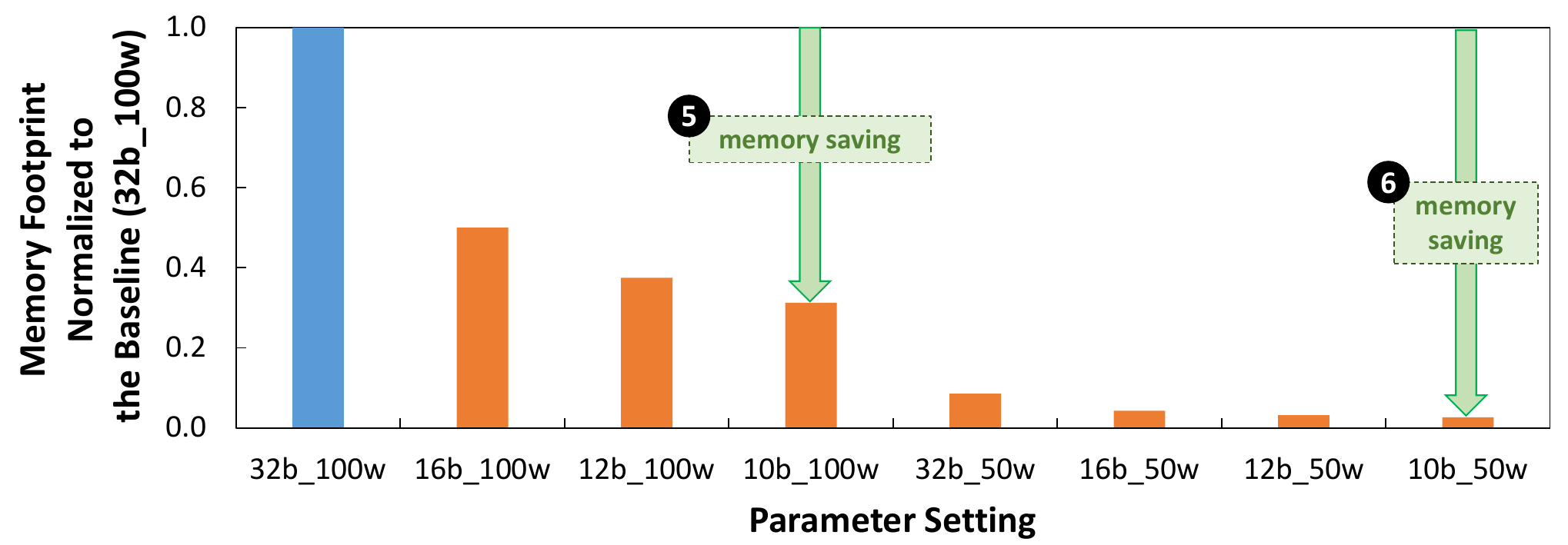}
\vspace{-0.2cm}
\caption{Experimental results for memory footprint normalized to the baseline (32b\_100w) across different precision levels and attention window sizes (i.e., 100$\times$100 and 50$\times$50).} 
\label{Fig_Result_Memory}
\vspace{-0.2cm}
\end{figure}

Experimental results for memory footprint are presented in Figure~\ref{Fig_Result_Memory}. 
These results show that, in general, our weight quantization step effectively reduces the memory footprint up to 68.75\% due to smaller precision levels for representing weight values. 
For instance, in the 100$\times$100 attention window case, our weight quantization leads to about 50\% memory saving for 16b\_100w, 62.50\% memory saving 12b\_100w, and 68.75\% memory saving 10b\_100w; see \circledB{5}. 
Meanwhile, in the 50$\times$50 attention window case, our weight quantization leads to about 91.42\% memory saving for 32b\_50w, 95.71\% memory saving for 16b\_50w, 96.78\% memory saving 12b\_50w, and 97.32\% memory saving 10b\_50w; see \circledB{6}.
The significant memory savings obtained in the 50$\times$50 attention window case come from the quantization as well as the reduction on the number of weights due to architectural differences in the network as shown in Table~\ref{Tab_SNNarch_100x100} and Table~\ref{Tab_SNNarch_50x50}. 

\vspace{-0.2cm}

\subsection{Processing Time Speed-Up}
\label{Sec_Res_Time}

\begin{figure}[t]
\centering
\includegraphics[width=0.8\linewidth]{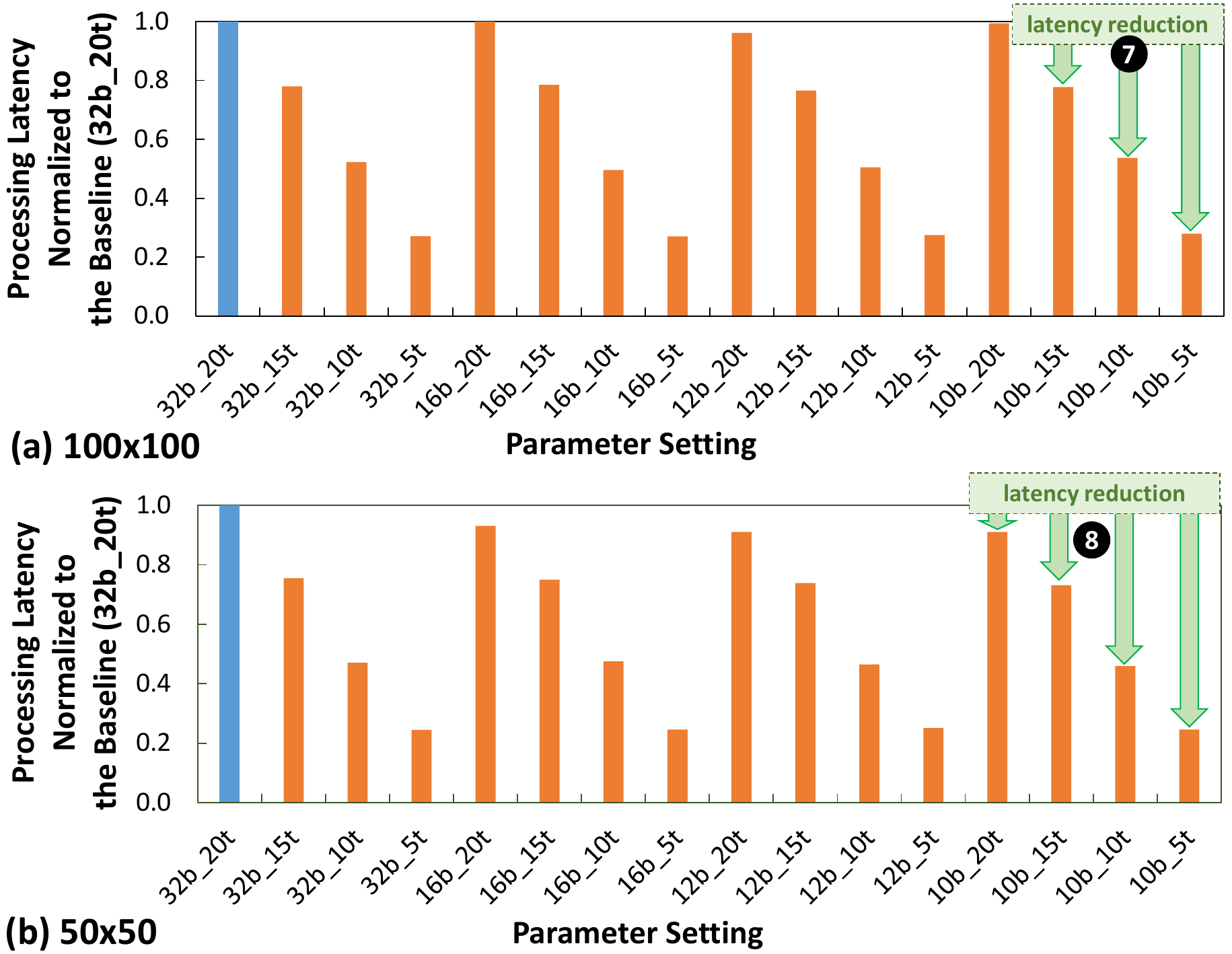}
\caption{Experimental results for latency normalized to the baseline (32b\_20t) across different parameter settings, including precision levels, timesteps, and training epochs under (a) 100$\times$100 attention window, and (b) 50$\times$50 attention window.} 
\label{Fig_Result_Latency}
\end{figure}

Experimental results for processing latency are presented in Figure~\ref{Fig_Result_Latency}.
These results show that, in general, our timestep reduction step effectively reduce the processing latency as compared to the baseline model (32b\_20t). 
For instance, in the 100$\times$100 attention window case, our timestep reduction leads to speed-ups by 1.27x - 1.31x for timestep=15 (i.e., 32b\_15t, 16b\_15t, 12b\_15t, and 10b\_15t); by 1.86x - 2.02x for timestep=10 (i.e., 32b\_10t, 16b\_10t, 12b\_10t, and 10b\_10t); as well as by 3.58x - 3.69x for timestep=5 (i.e., 32b\_5t, 16b\_5t, 12b\_5t, and 10b\_5t); as shown by \circledB{7}. 
Meanwhile, in the 50$\times$50 attention window case, our timestep reduction leads to speed-ups by 1.28x - 1.32x for timestep=15 (i.e., 32b\_15t, 16b\_15t, 12b\_15t, and 10b\_15t); by 2.04x - 2.11x for timestep=10 (i.e., 32b\_10t, 16b\_10t, 12b\_10t, and 10b\_10t); as well as by 3.85x - 3.95x for timestep=5 (i.e., 32b\_5t, 16b\_5t, 12b\_5t, and 10b\_5t); as shown by \circledB{8}. 
These significant speed-ups come from the reduced timesteps which curtail the processing time for spike trains. 

\subsection{Energy Efficiency Improvements}
\label{Sec_Res_Energy}

\begin{figure}[hbtp]
\centering
\includegraphics[width=0.8\linewidth]{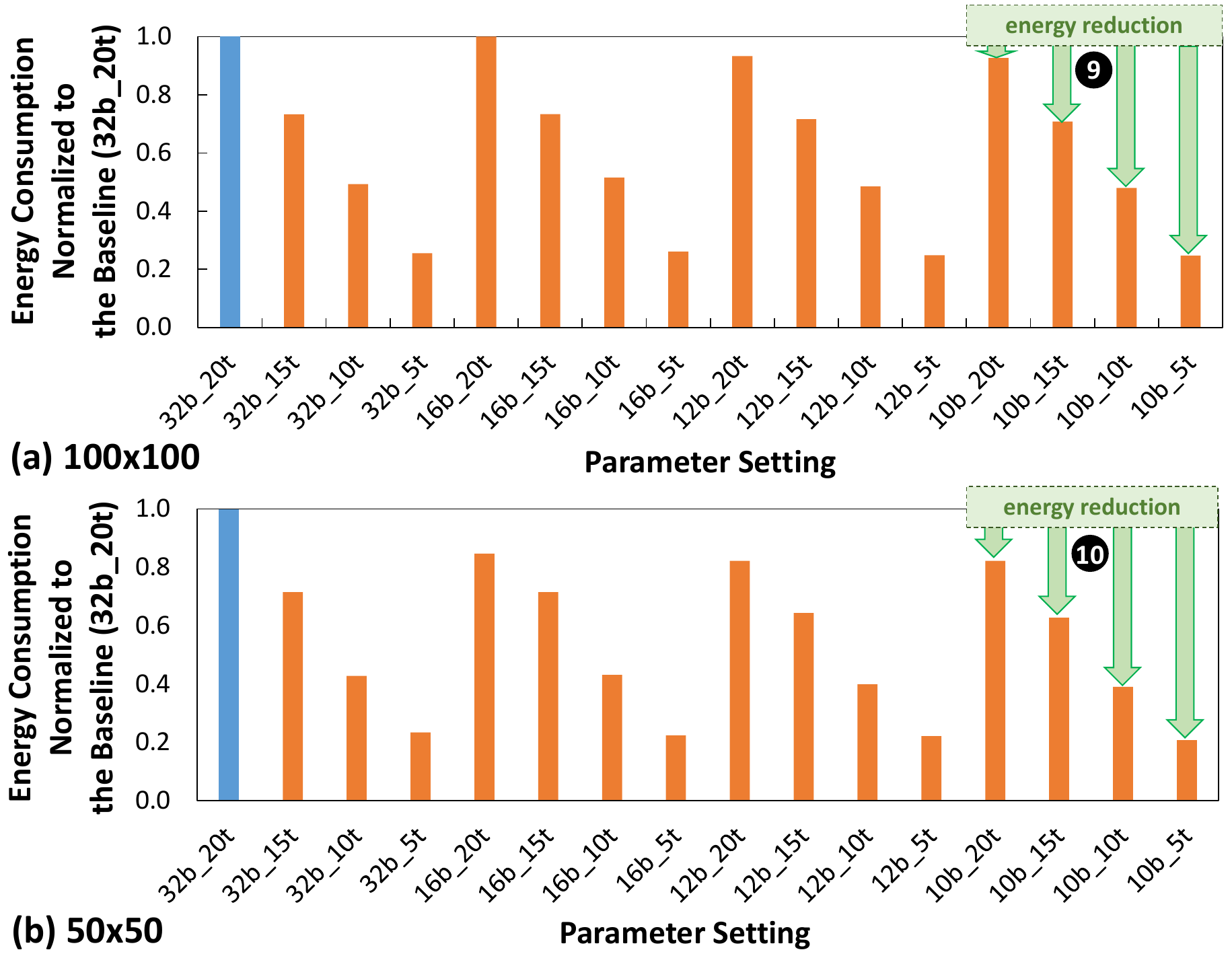}
\caption{Experimental results for energy consumption normalized to the baseline (32b\_20t) across different parameter settings, including precision levels, timesteps, and training epochs under (a) 100$\times$100 attention window, and (b) 50$\times$50 attention window.} 
\label{Fig_Result_Energy}
\vspace{-0.3cm}
\end{figure}

Experimental results for energy consumption are presented in Figure~\ref{Fig_Result_Energy}.
These results show that, in general, our optimization techniques effectively reduce the energy consumption as compared to the baseline model (32b\_20t). 
For instance, in the 100$\times$100 attention window case, our optimizations lead to energy improvements by 1.36x - 1.41x for settings 16b\_15t, 12b\_15t, and 10b\_15t; by 1.94x - 2.08x for settings 16b\_10t, 12b\_10t and 10b\_10t; as well as by 3.82x - 4.03x for settings 16b\_5t, 12b\_5t, and 10b\_5t; as shown by \circledB{9}. 
Meanwhile, in the 50$\times$50 attention window case, our optimizations lead to energy improvements by 1.35x - 1.54x for settings 16b\_15t, 12b\_15t, and 10b\_15t; by 2.24x - 2.48x for settings 16b\_10t, 12b\_10t and 10b\_10t; as well as by 4.32x - 4.66x for settings 16b\_5t, 12b\_5t, and 10b\_5t; as shown by \circledB{10}. 
These significant energy efficiency improvements come from the joint benefits from weight quantization, timestep reduction, and attention window reduction that decrease the processing power and processing latency, and hence the energy consumption.  

\subsection{Trade-Off Analysis}
\label{Sec_Res_TradeOff}

\begin{figure}[t]
\centering
\includegraphics[width=0.92\linewidth]{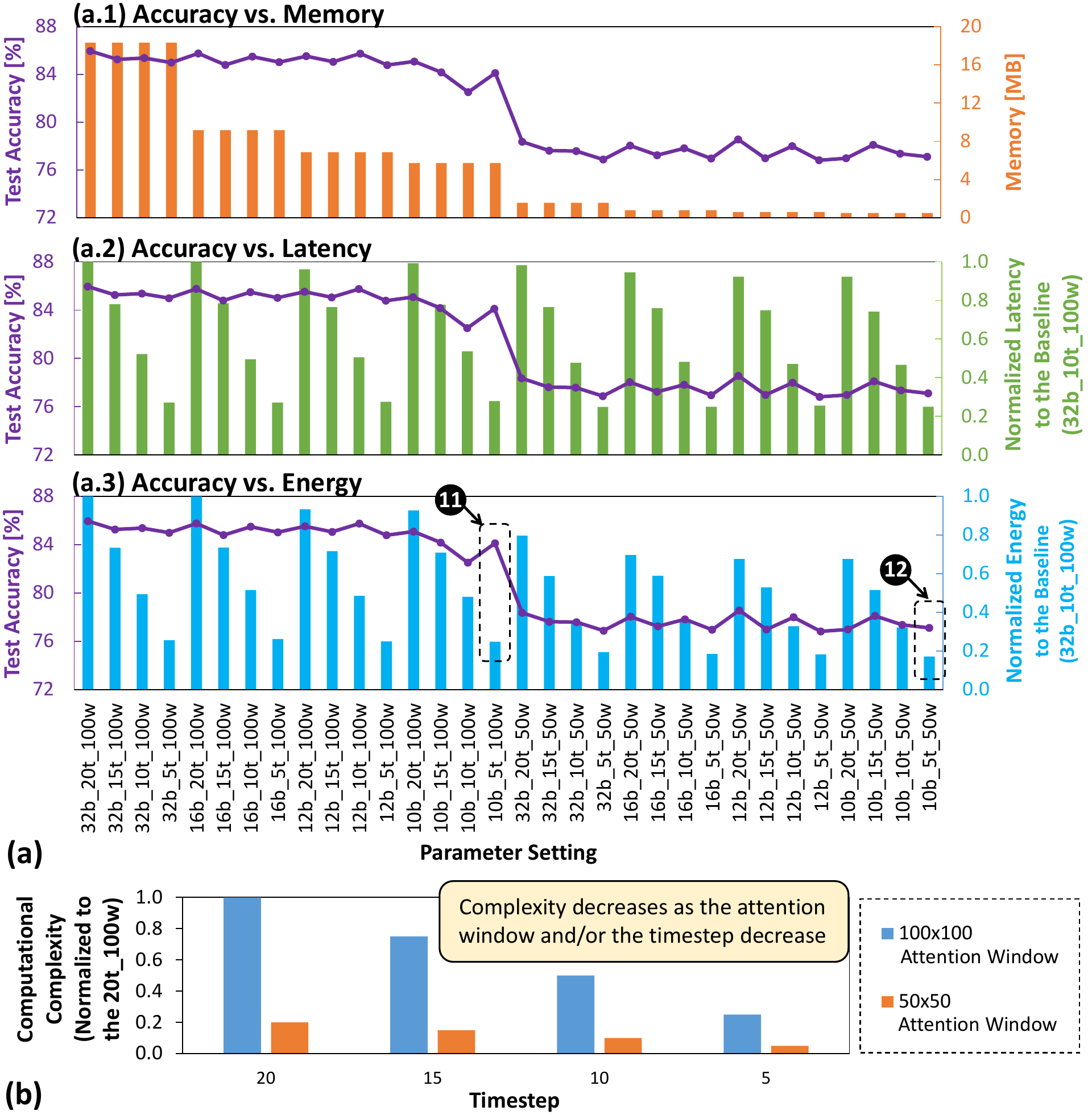}
\caption{(a) Trade-off analysis for accuracy vs. memory, accuracy vs. normalized latency, and accuracy vs. normalized energy consumption. 
(b) Computational complexity of different designs across different attention window sizes (i.e., 100$\times$100 and 50$\times$50) and timestep settings (i.e., 20, 15, 10, and 5).}
\label{Fig_Result_TradeOff}
\vspace{-0.4cm}
\end{figure}

To properly select the appropriate SNN model for the given memory and latency constraints, we perform a trade-off analysis. 
To do this, we analyze the correlation between accuracy and memory footprint, accuracy and latency, as well as accuracy and energy consumption.  
\textit{For the accuracy-memory analysis}, we observe that accuracy decreases as the memory footprint decreases, which is indicated by lower bit precision levels with smaller attention window sizes; see Figure~\ref{Fig_Result_TradeOff}(a.1). 
It means that we need to select the appropriate network model whose memory footprint meets the given memory constraint. 
For instance, if the memory constraint is 8MB, then we can select network models with 12- and 10-bit precision under 100$\times$100 attention window, as well as all network models under 50$\times$50 attention window. 
\textit{For the accuracy-latency analysis}, we observe that accuracy decreases as the latency decreases, which is indicated by shorter timesteps; see Figure~\ref{Fig_Result_TradeOff}(a.2). 
It means that we need to select the appropriate network model whose latency meets the given memory constraint. 
For instance, if the latency constraint is 0.25x from the baseline latency, then we can select network models with 5 timestep under both 100$\times$100 and 50$\times$50 attention window sizes. 
\textit{For the accuracy-energy analysis}, we observe that accuracy also decreases as the energy consumption decreases, which is indicated by lower bit precision levels with shorter timesteps and smaller attention window sizes; see Figure~\ref{Fig_Result_TradeOff}(a.3). 
For instance, if the given constraints are 8MB memory with 0.25x latency from the baseline latency, then we can select the network model with 10b\_5t\_100w setting, which achieves 84.12\% with 68.75\% memory saving and 3.58x speed-up, and 4.03x energy efficiency improvement from the baseline model; see label-\circledB{11}.
Furthermore, if the memory constraint is significantly smaller (e.g., 1MB memory) with 0.25x latency from the baseline latency, then we can select the network model with 10b\_5t\_50w setting, which achieves 77.10\% with 97.32\% memory saving and 4x speed-up, and 5.85x energy efficiency improvement from the baseline model; see label-\circledB{12}.

In summary, \textit{all these experimental results show that our SNN4Agents framework effectively improves energy efficiency of SNN models for autonomous agent applications}.
Furthermore, our framework also \textit{enables the users to find and select the suitable SNN model to meet the given memory and latency constraints}, i.e., by tuning the optimization settings for weight quantization, timestep setting, and attention window size.

\subsection{Computational Complexity}
\label{Sec_Res_Complexity}

Besides performance and efficiency benefits, a set of optimization techniques in our SNN4Agents framework also leads to different levels of computational complexity, which can be quantified through the number of synaptic and neuronal operations across different layers and the given timestep setting. 
Figure~\ref{Fig_Result_TradeOff}(b) shows the comparison of computational complexity from different designs across different attention window sizes (i.e., 100$\times$100 and 50$\times$50) and timestep settings (i.e., 20, 15, 10, and 5). 
In general, a smaller attention window incurs lower computational complexity than the bigger ones. 
The reason is that, a smaller attention window leads to smaller model size as indicated in Table~\ref{Tab_SNNarch_100x100} and Table~\ref{Tab_SNNarch_50x50}, and hence smaller number of synaptic and neuronal operations. 
Specifically, if the larger attention window size is $W_0 \times W_0$ and the smaller attention window size is $W_1 \times W_1$, then the reduction factor of the computational complexity achieved by employing the smaller attention window can be estimated with $\approx(W_0/W_1)^2$.
For instance, the 20t\_50w model has around 80\% lower computation than the state-of-the-art (i.e., 20t\_100w).
Furthermore, a smaller number of timestep setting also incurs lower computational complexity than the bigger ones, as the synaptic and neuronal operations are performed in a timestep unit. 
Specifically, if the larger timestep is $T_0$ and the smaller timestep is $T_1$, then the reduction factor of the computational complexity achieved by employing the smaller timestep can be computed with $(T_0/T_1)$.
For instance, the 15t\_100w, 10t\_100w, and 5t\_100w models incur lower computation than the 20t\_100w by 25\%, 50\%, and 75\%, respectively. 
Consequently, jointly employing smaller attention window size and smaller number of timestep will further decrease the computational complexity, whose reduction factor can be estimated with $\approx(W_0/W_1)^2 \cdot (T_0/T_1)$.  
For instance, the 15t\_50w, 10t\_50w, and 5t\_50w models incur lower computation than the 20t\_100w by about 85\%, 90\%, and 95\%, respectively.

\subsection{Further Discussion}
\label{Sec_Res_FurtherDisc}

Our SNN4Agents framework is the first work that incorporates a set of optimization techniques while considering the event-based automotive dataset for enabling the efficient development of SNN-based autonomous agents. Therefore, it has several advantages as the following.
\begin{itemize}
    \item Our framework incorporates each optimization technique in a modular form. Therefore, the existing optimization techniques in the framework can be activated or deactivated as per the design requirements. Furthermore, new optimization techniques can also be incorporated feasibly into the framework.
    \item Our framework can be used for design space exploration to investigate and understand the role of different SNN parameters. For instance, this framework can be utilized to observe the impact of specific parameters (e.g., membrane threshold potential) on the accuracy~\citep{Ref_Bano_SNNParameters_arXiv24}.
    \item The generated SNN model has direct interfacing with event-based input data stream, thereby enabling efficient integration with event-based sensors (e.g., DVS Camera). 
\end{itemize}
Besides these advantages, our framework in the current form can still be improved further to enhance the generated SNN model. Our SNN4Agents currently supports the STBP learning rule with offline-based training scenarios, which may be insufficient for some application use-cases. For instance, some autonomous agents may need to continuously adapt to dynamic operational environments, hence our SNN4Agents framework needs to be enhanced with advanced neural architectures and/or continual learning algorithms, while considering both offline and online training scenarios~\citep{Ref_Putra_EmbodiedNeuroAI4Robotics_arXiv24}.
Toward this, in the future, we plan to continue extending the work by incorporating more complex datasets in our framework, such as EventKITTI~\citep{Ref_Liang_EventKITTI_MFI22}, then evaluating the performance as well as testing it for a real-world robotic application use-case (e.g., UGV rover).

\section{Conclusion}
\label{Sec_Conclusion}

In this work, we propose an SNN4Agents framework that employs a set of optimization techniques for developing energy-efficient SNNs targeting autonomous agent applications. 
Here, our SNN4Agents compresses the SNN model through weight quantization, optimizes processing time through timestep reduction, and optimizes input samples through attention window reduction. 
The experimental results show that our proposed framework effectively improves energy efficiency, reduces memory footprint and latency, while maintaining high accuracy. 
If we consider a tolerance range of 2\% lower accuracy from the baseline, we can achieve 84.12\% accuracy with 68.75\% memory saving, 3.58x speed-up, and 4.03x energy efficiency improvement. 
In this manner, our SNN4Agents framework paves the way for further research and studies toward enabling the efficient development of SNN-based autonomous agents, and can be enhanced by incorporating other optimization techniques. 

\section*{Data Availability Statement}
\label{Sec_DataStatement}

Publicly available NCARS dataset can be accessed at https://www.prophesee.ai/2018/03/13/dataset-n-cars/. 
Meanwhile, codes of SNN4Agents are available at https://github.com/rachmadvwp/SNN4Agents.


\section*{Author Contributions}
\label{Sec_Author}

Rachmad Vidya Wicaksana Putra, Alberto Marchisio, and Muhammad Shafique contributed to the conception of the framework. 
Rachmad Vidya Wicaksana Putra and Alberto Marchisio developed and implemented the framework, as well as wrote the first draft of the manuscript.
Rachmad Vidya Wicaksana Putra organized the experiments.
Muhammad Shafique supervised the research work. 
All authors contributed to manuscript writing, read, and approved the submitted version.

\section*{Acknowledgments}
\label{Sec_Ack}

This work was partially supported by the NYUAD Center for Artificial Intelligence and Robotics (CAIR), funded by Tamkeen under the NYUAD Research Institute Award CG010. 



\bibliographystyle{frontiersinSCNS_ENG_HUMS} 
\bibliography{Bibliography}

\begin{thebibliography}{39}
\providecommand{\natexlab}[1]{#1}
\expandafter\ifx\csname urlstyle\endcsname\relax
  \providecommand{\doi}[1]{doi:\discretionary{}{}{}#1}\else
  \providecommand{\doi}{doi:\discretionary{}{}{}\begingroup
  \urlstyle{rm}\Url}\fi
\providecommand{\selectlanguage}[1]{\relax}
\providecommand{\bibAnnoteFile}[1]{%
  \IfFileExists{#1}{\begin{quotation}\noindent\textsc{Key:} #1\\
  \textsc{Annotation:}\ \input{#1}\end{quotation}}{}}
\providecommand{\bibAnnote}[2]{%
  \begin{quotation}\noindent\textsc{Key:} #1\\
  \textsc{Annotation:}\ #2\end{quotation}}

\bibitem[{Bano et~al.(2024)Bano, Putra, Marchisio, and
  Shafique}]{Ref_Bano_SNNParameters_arXiv24}
Bano, I., Putra, R. V.~W., Marchisio, A., and Shafique, M. (2024).
\newblock A methodology to study the impact of spiking neural network
  parameters considering event-based automotive data.
\newblock \emph{arXiv preprint arXiv:2404.03493}
\bibAnnoteFile{Ref_Bano_SNNParameters_arXiv24}

\bibitem[{Bartolozzi et~al.(2022)Bartolozzi, Indiveri, and
  Donati}]{Ref_Bartolozzi_EmbodiedNeuroIntel_Nature22}
Bartolozzi, C., Indiveri, G., and Donati, E. (2022).
\newblock Embodied neuromorphic intelligence.
\newblock \emph{Nature communications} 13, 1024
\bibAnnoteFile{Ref_Bartolozzi_EmbodiedNeuroIntel_Nature22}

\bibitem[{Bonnevie et~al.(2021)Bonnevie, Duberg, and
  Jensfelt}]{Ref_Bonnevie_DynamicEnv_ICARA21}
Bonnevie, R., Duberg, D., and Jensfelt, P. (2021).
\newblock Long-term exploration in unknown dynamic environments.
\newblock In \emph{2021 7th International Conference on Automation, Robotics
  and Applications (ICARA)}. 32--37.
\newblock \doi{10.1109/ICARA51699.2021.9376367}
\bibAnnoteFile{Ref_Bonnevie_DynamicEnv_ICARA21}

\bibitem[{Bu et~al.(2022)Bu, Fang, Ding, DAI, Yu, and
  Huang}]{Ref_Bu_ANNtoSNNConversion_ICLR22}
Bu, T., Fang, W., Ding, J., DAI, P., Yu, Z., and Huang, T. (2022).
\newblock Optimal {ANN}-{SNN} conversion for high-accuracy and
  ultra-low-latency spiking neural networks.
\newblock In \emph{International Conference on Learning Representations}
\bibAnnoteFile{Ref_Bu_ANNtoSNNConversion_ICLR22}

\bibitem[{Chowdhury et~al.(2021)Chowdhury, Rathi, and
  Roy}]{Ref_Chowdhury_OneTimestepIsAllYouNeed_arxiv21}
Chowdhury, S.~S., Rathi, N., and Roy, K. (2021).
\newblock One timestep is all you need: Training spiking neural networks with
  ultra low latency.
\newblock \emph{CoRR} abs/2110.05929
\bibAnnoteFile{Ref_Chowdhury_OneTimestepIsAllYouNeed_arxiv21}

\bibitem[{Cordone et~al.(2022)Cordone, Miramond, and
  Thierion}]{Ref_Cordone_ObjDetSNN_IJCNN22}
Cordone, L., Miramond, B., and Thierion, P. (2022).
\newblock Object detection with spiking neural networks on automotive event
  data.
\newblock In \emph{2022 International Joint Conference on Neural Networks
  (IJCNN)}. 1--8.
\newblock \doi{10.1109/IJCNN55064.2022.9892618}
\bibAnnoteFile{Ref_Cordone_ObjDetSNN_IJCNN22}

\bibitem[{Guo et~al.(2021)Guo, Fouda, Eltawil, and
  Salama}]{Ref_Guo_NeuralCoding_FNINS21}
Guo, W., Fouda, M.~E., Eltawil, A.~M., and Salama, K.~N. (2021).
\newblock Neural coding in spiking neural networks: \uppercase{A} comparative
  study for robust neuromorphic systems.
\newblock \emph{Frontiers in Neuroscience (FNINS)} 15.
\newblock \doi{10.3389/fnins.2021.638474}
\bibAnnoteFile{Ref_Guo_NeuralCoding_FNINS21}

\bibitem[{Gupta et~al.(2015)Gupta, Agrawal, Gopalakrishnan, and
  Narayanan}]{Ref_Gupta_Precision_ICML15}
Gupta, S., Agrawal, A., Gopalakrishnan, K., and Narayanan, P. (2015).
\newblock Deep learning with limited numerical precision.
\newblock In \emph{International Conference on Machine Learning (ICML)}, eds.
  F.~Bach and D.~Blei (Lille, France: PMLR), vol.~37 of \emph{Proceedings of
  Machine Learning Research}, 1737--1746
\bibAnnoteFile{Ref_Gupta_Precision_ICML15}

\bibitem[{Hao et~al.(2023)Hao, Ding, Bu, Huang, and
  Yu}]{Ref_Hao_ANNtoSNNCalibration_ICLR23}
Hao, Z., Ding, J., Bu, T., Huang, T., and Yu, Z. (2023).
\newblock Bridging the gap between {ANN}s and {SNN}s by calibrating offset
  spikes.
\newblock In \emph{The Eleventh International Conference on Learning
  Representations}
\bibAnnoteFile{Ref_Hao_ANNtoSNNCalibration_ICLR23}

\bibitem[{{Izhikevich}(2004)}]{Ref_Izhikevich_CompareModels_TNN04}
{Izhikevich}, E.~M. (2004).
\newblock Which model to use for cortical spiking neurons?
\newblock \emph{IEEE Transactions on Neural Networks (TNN)} 15, 1063--1070
\bibAnnoteFile{Ref_Izhikevich_CompareModels_TNN04}

\bibitem[{Li et~al.(2022)Li, He, Dong, Kong, and
  Zeng}]{Ref_Li_SpiCalib_arXiv22}
Li, Y., He, X., Dong, Y., Kong, Q., and Zeng, Y. (2022).
\newblock Spike calibration: Fast and accurate conversion of spiking neural
  network for object detection and segmentation.
\newblock \emph{arXiv preprint arXiv:2207.02702}
\bibAnnoteFile{Ref_Li_SpiCalib_arXiv22}

\bibitem[{Liang et~al.(2022)Liang, Cao, Yang, Zhang, and
  Chen}]{Ref_Liang_EventKITTI_MFI22}
Liang, Z., Cao, H., Yang, C., Zhang, Z., and Chen, G. (2022).
\newblock Global-local feature aggregation for event-based object detection on
  eventkitti.
\newblock In \emph{2022 IEEE International Conference on Multisensor Fusion and
  Integration for Intelligent Systems (MFI)}. 1--7.
\newblock \doi{10.1109/MFI55806.2022.9913852}
\bibAnnoteFile{Ref_Liang_EventKITTI_MFI22}

\bibitem[{Maass(1997)}]{Ref_Maass_SNN_NeuNet97}
Maass, W. (1997).
\newblock Networks of spiking neurons: The third generation of neural network
  models.
\newblock \emph{Neural Networks} 10, 1659--1671
\bibAnnoteFile{Ref_Maass_SNN_NeuNet97}

\bibitem[{Massa et~al.(2020)Massa, Marchisio, Martina, and
  Shafique}]{Massa_2020IJCNN_EfficientSNNGestures}
Massa, R., Marchisio, A., Martina, M., and Shafique, M. (2020).
\newblock An efficient spiking neural network for recognizing gestures with a
  {DVS} camera on the loihi neuromorphic processor.
\newblock In \emph{International Joint Conference on Neural Networks (IJCNN)}
  ({IEEE}), 1--9.
\newblock \doi{10.1109/IJCNN48605.2020.9207109}
\bibAnnoteFile{Massa_2020IJCNN_EfficientSNNGestures}

\bibitem[{Micikevicius et~al.(2018)Micikevicius, Narang, Alben, Diamos, Elsen,
  Garcia et~al.}]{Ref_Micikevicius_Precision_ICLR18}
Micikevicius, P., Narang, S., Alben, J., Diamos, G., Elsen, E., Garcia, D.,
  et~al. (2018).
\newblock Mixed precision training.
\newblock In \emph{International Conference on Learning Representation (ICLR)}
\bibAnnoteFile{Ref_Micikevicius_Precision_ICLR18}

\bibitem[{Neftci et~al.(2019)Neftci, Mostafa, and
  Zenke}]{Ref_Neftci_SurrogateSNNs_IEEEMSP19}
Neftci, E.~O., Mostafa, H., and Zenke, F. (2019).
\newblock Surrogate gradient learning in spiking neural networks: Bringing the
  power of gradient-based optimization to spiking neural networks.
\newblock \emph{IEEE Signal Processing Magazine} 36, 51--63
\bibAnnoteFile{Ref_Neftci_SurrogateSNNs_IEEEMSP19}

\bibitem[{Posch et~al.(2011)Posch, Matolin, and
  Wohlgenannt}]{Posch_2011JSSC_ATISSensor}
Posch, C., Matolin, D., and Wohlgenannt, R. (2011).
\newblock A {QVGA} 143 db dynamic range frame-free {PWM} image sensor with
  lossless pixel-level video compression and time-domain {CDS}.
\newblock \emph{{IEEE} J. Solid State Circuits} 46, 259--275.
\newblock \doi{10.1109/JSSC.2010.2085952}
\bibAnnoteFile{Posch_2011JSSC_ATISSensor}

\bibitem[{Putra et~al.(2021{\natexlab{a}})Putra, Hanif, and
  Shafique}]{Ref_Putra_ReSpawn_ICCAD21}
Putra, R. V.~W., Hanif, M.~A., and Shafique, M. (2021{\natexlab{a}}).
\newblock Respawn: Energy-efficient fault-tolerance for spiking neural networks
  considering unreliable memories.
\newblock In \emph{2021 IEEE/ACM International Conference On Computer Aided
  Design (ICCAD)}. 1--9.
\newblock \doi{10.1109/ICCAD51958.2021.9643524}
\bibAnnoteFile{Ref_Putra_ReSpawn_ICCAD21}

\bibitem[{Putra et~al.(2021{\natexlab{b}})Putra, Hanif, and
  Shafique}]{Ref_Putra_SparkXD_DAC21}
Putra, R. V.~W., Hanif, M.~A., and Shafique, M. (2021{\natexlab{b}}).
\newblock Sparkxd: A framework for resilient and energy-efficient spiking
  neural network inference using approximate dram.
\newblock In \emph{2021 58th ACM/IEEE Design Automation Conference (DAC)}.
  379--384.
\newblock \doi{10.1109/DAC18074.2021.9586332}
\bibAnnoteFile{Ref_Putra_SparkXD_DAC21}

\bibitem[{Putra et~al.(2022{\natexlab{a}})Putra, Hanif, and
  Shafique}]{Ref_Putra_EnforceSNN_FNINS22}
Putra, R. V.~W., Hanif, M.~A., and Shafique, M. (2022{\natexlab{a}}).
\newblock Enforcesnn: Enabling resilient and energy-efficient spiking neural
  network inference considering approximate drams for embedded systems.
\newblock \emph{Frontiers in Neuroscience (FNINS)} 16, 937782
\bibAnnoteFile{Ref_Putra_EnforceSNN_FNINS22}

\bibitem[{Putra et~al.(2022{\natexlab{b}})Putra, Hanif, and
  Shafique}]{Ref_Putra_SoftSNN_DAC22}
Putra, R. V.~W., Hanif, M.~A., and Shafique, M. (2022{\natexlab{b}}).
\newblock Softsnn: Low-cost fault tolerance for spiking neural network
  accelerators under soft errors.
\newblock In \emph{59th ACM/IEEE Design Automation Conference (DAC)}. 151--156
\bibAnnoteFile{Ref_Putra_SoftSNN_DAC22}

\bibitem[{Putra et~al.(2023)Putra, Hanif, and
  Shafique}]{Ref_Putra_RescueSNN_FNINS23}
Putra, R. V.~W., Hanif, M.~A., and Shafique, M. (2023).
\newblock Rescuesnn: enabling reliable executions on spiking neural network
  accelerators under permanent faults.
\newblock \emph{Frontiers in Neuroscience (FNINS)} 17.
\newblock \doi{10.3389/fnins.2023.1159440}
\bibAnnoteFile{Ref_Putra_RescueSNN_FNINS23}

\bibitem[{Putra et~al.(2024)Putra, Marchisio, Zayer, Dias, and
  Shafique}]{Ref_Putra_EmbodiedNeuroAI4Robotics_arXiv24}
Putra, R. V.~W., Marchisio, A., Zayer, F., Dias, J., and Shafique, M. (2024).
\newblock Embodied neuromorphic artificial intelligence for robotics:
  Perspectives, challenges, and research development stack.
\newblock \emph{arXiv preprint arXiv:2404.03325}
  \doi{10.48550/arXiv.2404.03325}
\bibAnnoteFile{Ref_Putra_EmbodiedNeuroAI4Robotics_arXiv24}

\bibitem[{{Putra} and {Shafique}(2020)}]{Ref_Putra_FSpiNN_TCAD20}
{Putra}, R. V.~W. and {Shafique}, M. (2020).
\newblock Fspinn: An optimization framework for memory-efficient and
  energy-efficient spiking neural networks.
\newblock \emph{IEEE Transactions on Computer-Aided Design of Integrated
  Circuits and Systems (TCAD)} 39, 3601--3613.
\newblock \doi{10.1109/TCAD.2020.3013049}
\bibAnnoteFile{Ref_Putra_FSpiNN_TCAD20}

\bibitem[{Putra and Shafique(2021{\natexlab{a}})}]{Ref_Putra_QSpiNN_IJCNN21}
Putra, R. V.~W. and Shafique, M. (2021{\natexlab{a}}).
\newblock Q-spinn: A framework for quantizing spiking neural networks.
\newblock In \emph{2021 International Joint Conference on Neural Networks
  (IJCNN)}. 1--8.
\newblock \doi{10.1109/IJCNN52387.2021.9534087}
\bibAnnoteFile{Ref_Putra_QSpiNN_IJCNN21}

\bibitem[{Putra and Shafique(2021{\natexlab{b}})}]{Ref_Putra_SpikeDyn_DAC21}
Putra, R. V.~W. and Shafique, M. (2021{\natexlab{b}}).
\newblock Spikedyn: A framework for energy-efficient spiking neural networks
  with continual and unsupervised learning capabilities in dynamic
  environments.
\newblock In \emph{58th ACM/IEEE Design Automation Conference (DAC)}.
  1057--1062
\bibAnnoteFile{Ref_Putra_SpikeDyn_DAC21}

\bibitem[{Putra and Shafique(2022)}]{Ref_Putra_lpSpikeCon_IJCNN22}
Putra, R. V.~W. and Shafique, M. (2022).
\newblock lpspikecon: Enabling low-precision spiking neural network processing
  for efficient unsupervised continual learning on autonomous agents.
\newblock In \emph{2022 International Joint Conference on Neural Networks
  (IJCNN)}. 1--8.
\newblock \doi{10.1109/IJCNN55064.2022.9892948}
\bibAnnoteFile{Ref_Putra_lpSpikeCon_IJCNN22}

\bibitem[{Putra and Shafique(2023{\natexlab{a}})}]{Ref_Putra_Mantis_ICARA23}
Putra, R. V.~W. and Shafique, M. (2023{\natexlab{a}}).
\newblock Mantis: Enabling energy-efficient autonomous mobile agents with
  spiking neural networks.
\newblock In \emph{2023 9th International Conference on Automation, Robotics
  and Applications (ICARA)}. 197--201.
\newblock \doi{10.1109/ICARA56516.2023.10125781}
\bibAnnoteFile{Ref_Putra_Mantis_ICARA23}

\bibitem[{Putra and Shafique(2023{\natexlab{b}})}]{Ref_Putra_TopSpark_IROS23}
Putra, R. V.~W. and Shafique, M. (2023{\natexlab{b}}).
\newblock Topspark: A timestep optimization methodology for energy-efficient
  spiking neural networks on autonomous mobile agents.
\newblock In \emph{2023 IEEE/RSJ International Conference on Intelligent Robots
  and Systems (IROS)}. 3561--3567.
\newblock \doi{10.1109/IROS55552.2023.10342499}
\bibAnnoteFile{Ref_Putra_TopSpark_IROS23}

\bibitem[{Putra and Shafique(2024)}]{Ref_Putra_SpikeNAS_arXiv24}
Putra, R. V.~W. and Shafique, M. (2024).
\newblock Spikenas: A fast memory-aware neural architecture search framework
  for spiking neural network-based autonomous agents.
\newblock \emph{arXiv preprint arXiv:2402.11322}
  \doi{10.48550/arXiv.2402.11322}
\bibAnnoteFile{Ref_Putra_SpikeNAS_arXiv24}

\bibitem[{Rathi et~al.(2023)Rathi, Chakraborty, Kosta, Sengupta, Ankit, Panda
  et~al.}]{Ref_Rathi_SNNsurvey_CSUR23}
Rathi, N., Chakraborty, I., Kosta, A., Sengupta, A., Ankit, A., Panda, P.,
  et~al. (2023).
\newblock Exploring neuromorphic computing based on spiking neural networks:
  Algorithms to hardware.
\newblock \emph{ACM Computing Survey} 55.
\newblock \doi{10.1145/3571155}
\bibAnnoteFile{Ref_Rathi_SNNsurvey_CSUR23}

\bibitem[{Roy et~al.(2019)Roy, Jaiswal, and
  Panda}]{Ref_Roy_SpikeMachineIntel_Nature19}
Roy, K., Jaiswal, A., and Panda, P. (2019).
\newblock Towards spike-based machine intelligence with neuromorphic computing.
\newblock \emph{Nature} 575, 607--617
\bibAnnoteFile{Ref_Roy_SpikeMachineIntel_Nature19}

\bibitem[{R{\"{u}}ckauer et~al.(2019)R{\"{u}}ckauer, K{\"{a}}nzig, Liu,
  Delbr{\"{u}}ck, and
  Sandamirskaya}]{Ruckauer_2019arxiv_NonDifferentiableLossFunctionSNN}
R{\"{u}}ckauer, B., K{\"{a}}nzig, N., Liu, S., Delbr{\"{u}}ck, T., and
  Sandamirskaya, Y. (2019).
\newblock Closing the accuracy gap in an event-based visual recognition task.
\newblock \emph{CoRR} abs/1906.08859
\bibAnnoteFile{Ruckauer_2019arxiv_NonDifferentiableLossFunctionSNN}

\bibitem[{Schuman et~al.(2022)Schuman, Kulkarni, Parsa, Mitchell, Kay
  et~al.}]{Ref_Schuman_OpportunityNeuro_Nature22}
Schuman, C.~D., Kulkarni, S.~R., Parsa, M., Mitchell, J.~P., Kay, B., et~al.
  (2022).
\newblock Opportunities for neuromorphic computing algorithms and applications.
\newblock \emph{Nature Computational Science} 2, 10--19
\bibAnnoteFile{Ref_Schuman_OpportunityNeuro_Nature22}

\bibitem[{Sironi et~al.(2018)Sironi, Brambilla, Bourdis, Lagorce, and
  Benosman}]{Ref_Sironi_HATS_CVPR18}
Sironi, A., Brambilla, M., Bourdis, N., Lagorce, X., and Benosman, R. (2018).
\newblock Hats: Histograms of averaged time surfaces for robust event-based
  object classification.
\newblock In \emph{2018 IEEE/CVF Conference on Computer Vision and Pattern
  Recognition (CVPR)}. 1731--1740.
\newblock \doi{10.1109/CVPR.2018.00186}
\bibAnnoteFile{Ref_Sironi_HATS_CVPR18}

\bibitem[{Viale et~al.(2021)Viale, Marchisio, Martina, Masera, and
  Shafique}]{Ref_Viale_CarSNN_IJCNN21}
Viale, A., Marchisio, A., Martina, M., Masera, G., and Shafique, M. (2021).
\newblock Carsnn: An efficient spiking neural network for event-based
  autonomous cars on the loihi neuromorphic research processor.
\newblock In \emph{2021 International Joint Conference on Neural Networks
  (IJCNN)}. 1--10.
\newblock \doi{10.1109/IJCNN52387.2021.9533738}
\bibAnnoteFile{Ref_Viale_CarSNN_IJCNN21}

\bibitem[{Viale et~al.(2022)Viale, Marchisio, Martina, Masera, and
  Shafique}]{Ref_Viale_LaneSNN_IROS22}
Viale, A., Marchisio, A., Martina, M., Masera, G., and Shafique, M. (2022).
\newblock Lanesnns: Spiking neural networks for lane detection on the loihi
  neuromorphic processor.
\newblock In \emph{2022 IEEE/RSJ International Conference on Intelligent Robots
  and Systems (IROS)} (IEEE), 79--86
\bibAnnoteFile{Ref_Viale_LaneSNN_IROS22}

\bibitem[{Wang et~al.(2014)Wang, Guo, and
  Adjouadi}]{Wang_2014EMBC_LeakyIntegrateAndFire}
Wang, Z., Guo, L., and Adjouadi, M. (2014).
\newblock A biological plausible generalized leaky integrate-and-fire neuron
  model.
\newblock In \emph{36th Annual International Conference of the {IEEE}
  Engineering in Medicine and Biology Society, {EMBC} 2014, Chicago, IL, USA,
  August 26-30, 2014} ({IEEE}), 6810--6813.
\newblock \doi{10.1109/EMBC.2014.6945192}
\bibAnnoteFile{Wang_2014EMBC_LeakyIntegrateAndFire}

\bibitem[{Wu et~al.(2018)Wu, Deng, Li, Zhu, and Shi}]{Ref_Wu_STBP_FNINS18}
Wu, Y., Deng, L., Li, G., Zhu, J., and Shi, L. (2018).
\newblock Spatio-temporal backpropagation for training high-performance spiking
  neural networks.
\newblock \emph{Frontiers in Neuroscience (FNINS)} 12, 331
\bibAnnoteFile{Ref_Wu_STBP_FNINS18}

\end{thebibliography}


\end{document}